\documentclass[journal]{IEEEtran}

\makeatletter
\def\endthebibliography{%
  \def\@noitemerr{\@latex@warning{Empty `thebibliography' environment}}%
  \endlist
}
\makeatother

\usepackage{color}
\usepackage{multirow}
\usepackage{adjustbox}
\usepackage[algo2e]{algorithm2e}
\usepackage{algorithm}
\usepackage{algorithmic}
\usepackage{graphicx}
\usepackage{epstopdf}
\usepackage{caption}
\usepackage{proof}
\usepackage{amsthm}
\usepackage{amsmath}
\DeclareCaptionType{copyrightbox}
\usepackage{subcaption}
\usepackage{setspace}
\usepackage{pbox}
\usepackage{soul}

\usepackage{amssymb}
\usepackage{wasysym}

\graphicspath{ {./figure/} }

\ifCLASSINFOpdf

\else

\fi

\begin{document}

\title{Repulsive Mixture Models of \\Exponential Family PCA for Clustering}

\author{Maoying~Qiao,
Tongliang~Liu,
Jun~Yu,~\IEEEmembership{Member,~IEEE,}
        Wei~Bian,~\IEEEmembership{Member,~IEEE,}
        and~Dacheng~Tao,~\IEEEmembership{Fellow,~IEEE}
\thanks{
M. Qiao and J. Yu are with the
School of Computer Science and Technology at Hangzhou Dianzi University, Hangzhou, 310018, China.
W. Bian is with the Centre for Artificial Intelligence and the Faculty of Engineering and Information Technology at the University of Technology Sydney, NSW 2007, Australia.
T. Liu and D. Tao are with UBTECH Sydney AI Centre and the School of Information Technologies in the Faculty of Engineering and Information Technologies at the University of Sydney, NSW 2006, Australia.
}
} 


\maketitle

\begin{abstract}
The mixture extension of exponential family principal component analysis (EPCA) was designed to encode much more structural information about data distribution than the traditional EPCA does. For example, due to the linearity of EPCA's essential form, nonlinear cluster structures cannot be easily handled, but they are explicitly modeled by the mixing extensions. However,  the traditional mixture of local EPCAs has the problem of model redundancy, i.e., overlaps among mixing components, which may cause ambiguity for data clustering. To alleviate this problem, in this paper, a repulsiveness-encouraging prior is introduced among mixing components and a diversified EPCA mixture (DEPCAM) model is developed in the Bayesian framework. Specifically, a determinantal point process (DPP) is exploited as a diversity-encouraging prior distribution over the joint local EPCAs. As required, a matrix-valued measure for L-ensemble kernel is designed, within which, $\ell_1$ constraints are imposed to facilitate selecting effective PCs of local EPCAs, and angular based similarity measure are proposed. An efficient variational EM algorithm is derived to perform parameter learning and hidden variable inference. Experimental results on both synthetic and real-world datasets confirm the effectiveness of the proposed method in terms of model parsimony and generalization ability on unseen test data.
\end{abstract}

\begin{IEEEkeywords}
Diversified model, Exponential Family PCA, Mixture models, Determinantal point processes.
\end{IEEEkeywords}

\IEEEpeerreviewmaketitle

\section{Introduction}
\label{sec:intro}

Principal component analysis (PCA) has been recognized as one of the most fundamental tools for data analysis. Assuming that data mainly lies on a compact subspace of the ambient feature space, it provides an efficient way to extract low-dimensional representations from high-dimensional data \cite{jolliffe2002principal} \cite{anzai2012pattern}.
In the past few decades, PCA was applied to a wide variety of problems, from image analysis \cite{turaga2002face}\cite{geladi1989principal}, image visualization \cite{kambhatla1997dimension}, to data compression \cite{du2007hyperspectral} and time series prediction \cite{ku1995disturbance}.

In order to encode prior information, recent studies have proposed important extensions of the conventional PCA, especially through the Bayesian method.
Bayesian PCA (BPCA) \cite{nounou2002bayesian} is a remarkable one, which reformulates PCA into a probabilistic framework by imposing a Gaussian latent variable model over the parameters of data representations.
{Exponential family PCA (EPCA)} \cite{collins2001generalization} extends BPCA by replacing the Gaussian distribution over latent variables with a more general form of distribution from the exponential family. Automatic model selections for BPCA and EPCA have also been studied, e.g., by using the Bayesian method \cite{minka2000automatic} and the automatic relevance determination (ARD) technique \cite{mackay1995probable}.


A ubiquitous phenomenon that cannot be modelled by PCA and above-mentioned variants is that datasets in practice often possess more structures, especially clusters, in data distribution. Such distributional structures are quite common, because particular subsets of a big dataset may have different semantic meanings or represent different visual concepts.
Previous work \cite{li2013simple} proposed a straightforward mixture extension of EPCA to incorporate the cluster structures of data distribution. However, the extension suffers from two limitations. Firstly, traditional mixture models may exhibit model redundancy and severe over-fitting problems. Because they treat their mixing components rather independently, the fitted components may overlap with each other. As a result, it may cause ambiguity when considering data separability. Also it may require more mixing components than it really needs to cover the whole observation space. Secondly, how to effectively determine the numbers of effective principal components (PCs) is still an important but not solved issue in practice for mixture models of EPCA.

In this paper, we propose a novel EPCA mixture model with diversity properties, i.e., diversified exponential family PCA mixture models (DEPCAM), which overcomes aforementioned limitations of previous extensions. Specifically, a diversity-encouraging prior, formulated with the determinantal point processes (DPP), is imposed over the joint set of the mixing components of local PCAs. It reduces the model redundancy by forcing the mixing components as far as possible in the model space, i.e., each mixing component to represent different part of the data space. Further, constraints are introduced to construct the diversity kernel of the DPP prior, which helps define the repulsiveness of the mixing components and provides an automatic scheme to determine the dimensions of the local PCAs of each mixing component. An expectation-maximization (EM) algorithm is derived for model parameter learning and hidden variable inference. The effectiveness of the proposed DEPCAM is verified on both synthetic and real datasets.

This paper is organized as follows.
Section \ref{sec:RWork} reviews related works including mixture models and diversity related topics.
Section \ref{sec:background} briefly introduces technical background.
Our model is developed in
Section \ref{sec:ourmodel}, and its parameter learning and inference are derived in
Section \ref{sec:learning}.
Finally, experimental results of demonstration and comparison are presented in
Section \ref{sec:exp}, and
Section \ref{sec:conclusion} concludes this paper.

\section{Related Works}
\label{sec:RWork}

\subsection{PCA and its Mixture Extensions}

PCA and its probabilistic extension {\cite{tipping1999probabilistic}} are popular dimensionality reduction tools for data analysis. They extract low-dimensional data representations via linear transformation. However they assume that the observations in high-dimensional space are from Gaussian distributions, which limits them to the continuous data space and fails them to handle more general-typed data such as integers. To address this limitation, Collins et al. \cite{collins2001generalization} proposes to replace the Gaussian function with exponential families and develops an EPCA. {Later on, extensions such as Bayesian EPCA \cite{li2013simple} with Bayesian inference have also been developed.}

{
Complex cluster structure is ubiquitous in data and needs to be identified, such as multiple character recognition \cite{kim2002numeral} and
multi-mode process monitoring \cite{xu2011multi}.
Previous studies have adapted the one component real-valued PCA to mixture models by mixing local PCA reconstructions based on identifying clusters.
On the one hand, mixing local data representations is a simple and straightforward way to do the mixing extension. Several works such as \cite{tipping1999mixtures}\cite{mahantesh2014study}\cite{zhang2004mixture}\cite{wang2005subspace}\cite{watanabe2009variational} have established such PCA mixture models by assuming that one single PCA is adequate for capturing each local cluster in the data representation space.
The components are combined via mixing coefficients. As a result, the nonlinear cluster property exhibiting in data is reserved.
On the other hand, there exist one disadvantage of such mixture extension. To handle the complexity in data, it increases the number of model parameters which might increase the risk of overfitting. In this paper, we alleviate such overfitting risk by imposing a diversity-encouraging prior.
}

Model selection has been an important issue for PCA mixture models, and has been explored by many researchers. Zhao \cite{zhao2014efficient} proposed a hierarchical Bayesian information criterion (BIC) to efficiently do model selection, where each BIC is penalized by its own effective sample size, rather than the larger whole sample size. Li and Tao \cite{li2013simple} applied ARD over latent transformation matrix variables to determine the effective number of PCs. Huang et al. \cite{huang2004minimum} explored a general notation of dimensionality in mixture models, and introduces a robust minimum effective dimension (MED) criterion to address the model selection issue. Kim et al. \cite{kim2001pca} proposed a fast and sub-optimal method for model order selection, and achieves this goal by pruning its insignificant PCA bases.

\subsection{Diversity}

Diversity is a naturally existing property and has been exploited in various application scenarios. For example, a news summary timeline may only describe the most important events from a large news corpus, but a diverse subset will be encouraged to cover this corpus as large as possible \cite{kulesza2011learning}. Another example is in an image retrieval scenario \cite{kulesza2011kDPP}. An image query may associate with ambiguous concepts, rather than a pure one. A diverse image subset returned by an image search engine will surely increase the chance to hit the intention of the query. One more example is in a video summarization task \cite{gong2014diverse}. A sequentially diverse frame subset can better extract an abstract representation of an action, rather than just extract defined important frames. In summary, diversity is a ubiquitous and important attribute in real world applications.

DPP \cite{Alex2012} has been introduced to formally model diversity in a fully probabilistic framework, and provide an effective way to integrate the diversity property as a prior into traditional probabilistic models. For example, Zou \cite{JamesZou2012} firstly applied the diversity formulated with DPPs as a prior over latent variables of the generative latent Dirichlet allocation (LDA) model. It enforces diversity over joint word-topic distributions to obtain well-spread topic models. From the view of LDA as a mixture model, each topic model can be seen as a mixing component, and the diversity-encouraging prior is placed over the topic distributions. Comparatively, the whole framework of the proposed method is quite similar to this work, but their specifications are totally different. Another remarkable work is a determinantal clustering process (DCP), proposed by Shah and Ghahramani \cite{shah2013determinantal}. This process places a diversity-opposite prior over all possible partitions of a dataset, which results in a nonparametric Bayesian approach for clustering tasks. More examples treating the diversity as a Bayesian prior include a spike-and-slab prior for variable selection in linear regression \cite{Mutsuki2014}, a diversity-encouraging prior over sequential neurons to capture and visualize their complex inhibitory and competitive relationships \cite{Jasper2013}, and diversified transition matrix for hidden Markov model \cite{qiao2015diversified}. We refer interested readers to \cite{xie2015diversifying}\cite{xie2015learning} \cite{XieZhuXing2016} \cite{batmanghelich2014diversifying} for more DPP-related model developments.

{For easy reference, PCA related acronyms are summarized in Table \ref{tab:acronyms}}.
\begin{table}[!h]
\centering
\begin{tabular}{|l|l|}
 \hline
 Acronyms & Full phrases \\ \hline
 PCA & Principal Component Analysis  \\
 BPCA & Bayesian PCA \\
 EPCA & Exponential family PCA   \\
 DEPCAM & Diversified EPCA Mixture models \\
 SePCA & Simple exponential family PCA  \\
 SePCA-MM & SePCA Mixture Models \\
 \hline
\end{tabular}
 \caption{Acronyms for PCA variants.}
\label{tab:acronyms}
\end{table}

\section{Background}
\label{sec:background}

\subsection{Simple Exponential Family PCA}

Simple exponential family PCA (SePCA) \cite{li2013simple}) has several advantages over the other PCA-related state-of-the-art methods.
It is more appropriate to handle more general-typed data, e.g., integers. It inherits all the advantages of Bayesian inference as under a Bayesian framework, for example, dealing with over-fitting problem and automatically determining the effective number of PCs. Therefore, we adopt SePCA as one of our basic building blocks.

Given the observable variables $X = \{x_n\}_{n=1}^N$ with $x_n\in \mathcal{Z}^D$ or $2^D$  and $D$ representing feature dimension, latent variables $Y=\{y_n\}_{n=1}^N$, and transformation matrix $W \in \mathcal{R}^{D\times d}$, an SePCA is modeled by a probability distribution over it. Specifically, it is formulated as
\begin{align}
p(X, Y, W; {\zeta}) = p(X|Y, W)p(Y)p(W; {\zeta}).
\label{eq:SePCA}
\end{align}
Here $y_n\in \mathcal{R}^d$ represents a low-dimensional representation of $d$ dimensional and $d \ll D$. Throughout the paper, we use the semicolon ``$;$" to separate the variables and parameters of a distribution, and the symbol ``$|$'' to help denote conditional distributions. For example, here $p(X,Y,W;{\zeta})$ represents a joint distribution of a variable set $\{X,Y,W\}$, which is parameterized by ${\zeta}$. Another example is that the expression $p(X|Y,W)$ stands for a conditional distribution of $X$ given the values of variables $Y,W$. A special case is when neither of the symbols appears in a distribution, like $p(Y)$. In this situation, the parameters of the distribution is omitted if no confusion arises.

We elaborate the three distributions on the right-hand side of the above equation one by one.
The first one is the likelihood of observations. Given both latent representations $Y$ and PCs $W$, the distribution for each observation $x_n$ is assumed to be independent with each other, and employs an exponential family.
Specifically, the likelihood formulation in the natural form of an exponential family, i.e.,
\begin{align}
p(x_n|y_n,W) = \exp \{x_n^\top W y_n + g(W y_n)+h(x_n) \},
\label{eq:SePCALikelihood}
\end{align}
is parametrized by its natural parameters $Wy_n$. Here, $\top$ denotes vector or matrix transpose operator, and
$g(\cdot)$ and $h(\cdot)$ are known functions given a specific member of the exponential family.
The second term on the right-hand side of (\ref{eq:SePCA}) represents a prior distribution over latent low-dimensional representations. It is assigned a zero-meaned, isotropic unit Gaussian distribution, i.e.,  $ y_n \sim \mathcal{N}(y_n|0,I)$.
The third term represents a prior over the transformation matrix $W$, which is composed by PCs.
As required by PCA, this transformation matrix should be orthogonal in order to retain the uncorrelated property amongst PCs.
Therefore, each column of the transformation matrix, i.e., each PC $w_i$, should be independent with each other, and is assigned a prior of an isotropic Gaussian distribution controlled by a precision hyper-parameter ${\zeta_i}$. Formally, the prior distribution over $W$ is given by
\begin{align}
p(W;{\zeta}) = \prod_{i=1}^d  \mathcal{N}(w_i;0, {\zeta_i^{-1}}\mathbf{I}),
\label{eq:singleWprior}
\end{align}
with $w_{i}$ denoting the $i$-th column of $W$ and $\mathcal{N}(w_i;0, {\zeta_i^{-1}}\mathbf{I})$ a Gaussian distribution over $w_i$ parameterized by mean parameter $0$ and precision parameter ${\zeta_i}$.
{The model selection of this formulation for effective number of PCs is motivated by ARD \cite{mackay1995probable}, and is implemented by
switching on or off each $w_i$ with learned} {$\zeta_i$ }
{according to the rule that
the $w_i$ is less significant if its corresponding} ${\zeta_i}$ is larger.

\subsection{SePCA Mixture Models}

There are many complicated situations where SePCA is incapable to handle but a mixture model of several SePCAs is competent.
For instance,
groups of observations in a high-dimensional space might not be separable any more after being transformed into a low-dimensional space by SePCA. This is caused by the definition of PCA whose projection axes are linear and dominated by the variance of all observations from all groups. One simple and effective way to increase the SePCA's model capacity in order to handle these cluster structures is to fit different groups of observations with different local SePCAs, rather than one holistic SePCA. Take a mixture extension of SePCA, i.e., SePCA Mixture Models (SePCA-MM) as an example. One mixing component of SePCA-MM is supposed to fit one cluster of observations, and then all mixing components are integrated via mixing coefficients. As a result, different groups of observations will be projected into different low-dimensional spaces, thus the separability exhibiting in the original observation space is preserved in several low-dimensional spaces. In conclusion, a mixture model is a natural choice for the above situation, and is able to address the modeling limitation of linearity of SePCA to some extent.

\begin{figure*}[!ht]
  \centering
  \begin{subfigure}[b]{0.49\textwidth}
  \includegraphics[width=1\textwidth]{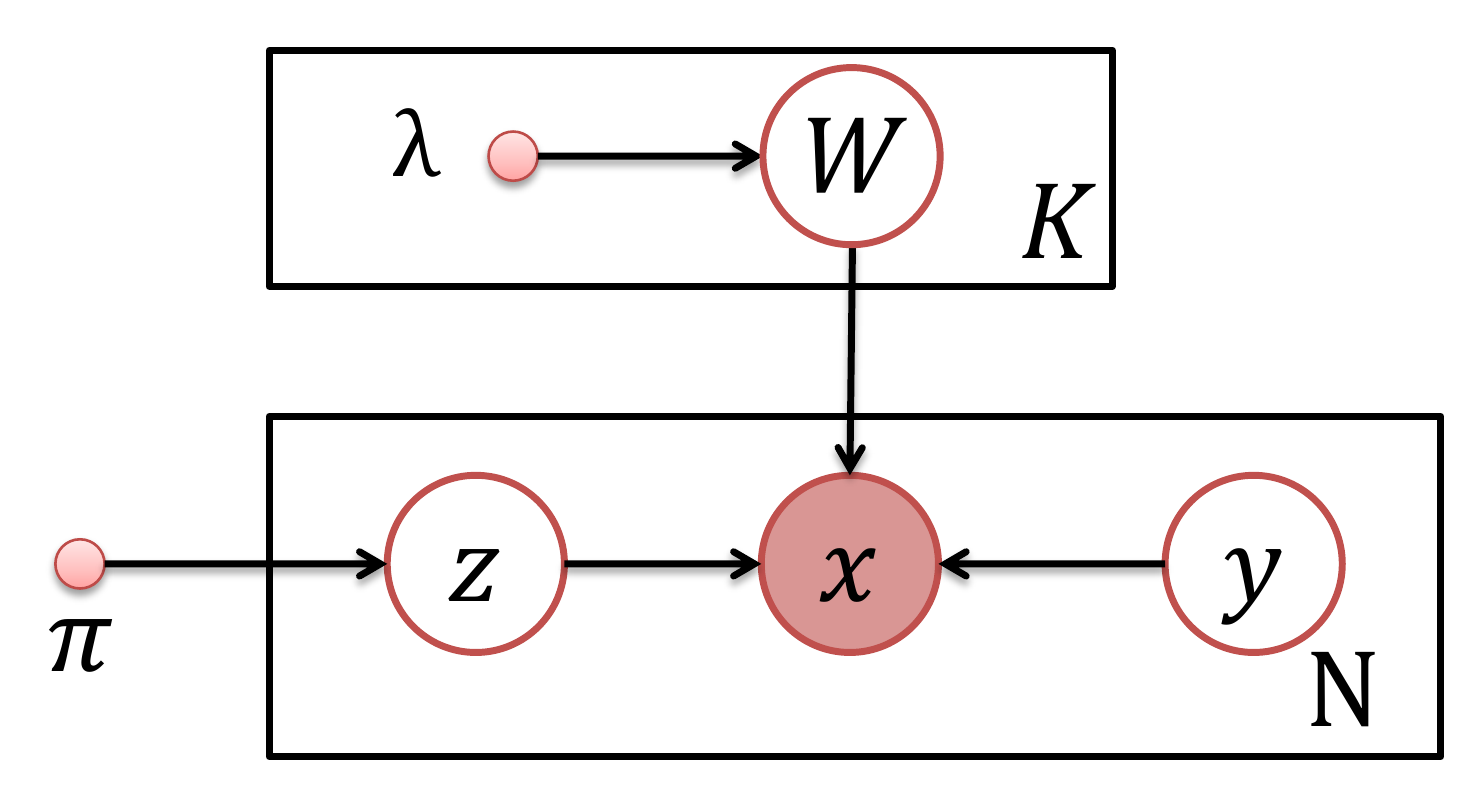}
  \caption{SePCA-MM}
  \label{fig:efpcamm}
  \end{subfigure}
  \begin{subfigure}[b]{0.49\textwidth}
  \includegraphics[width=1\textwidth]{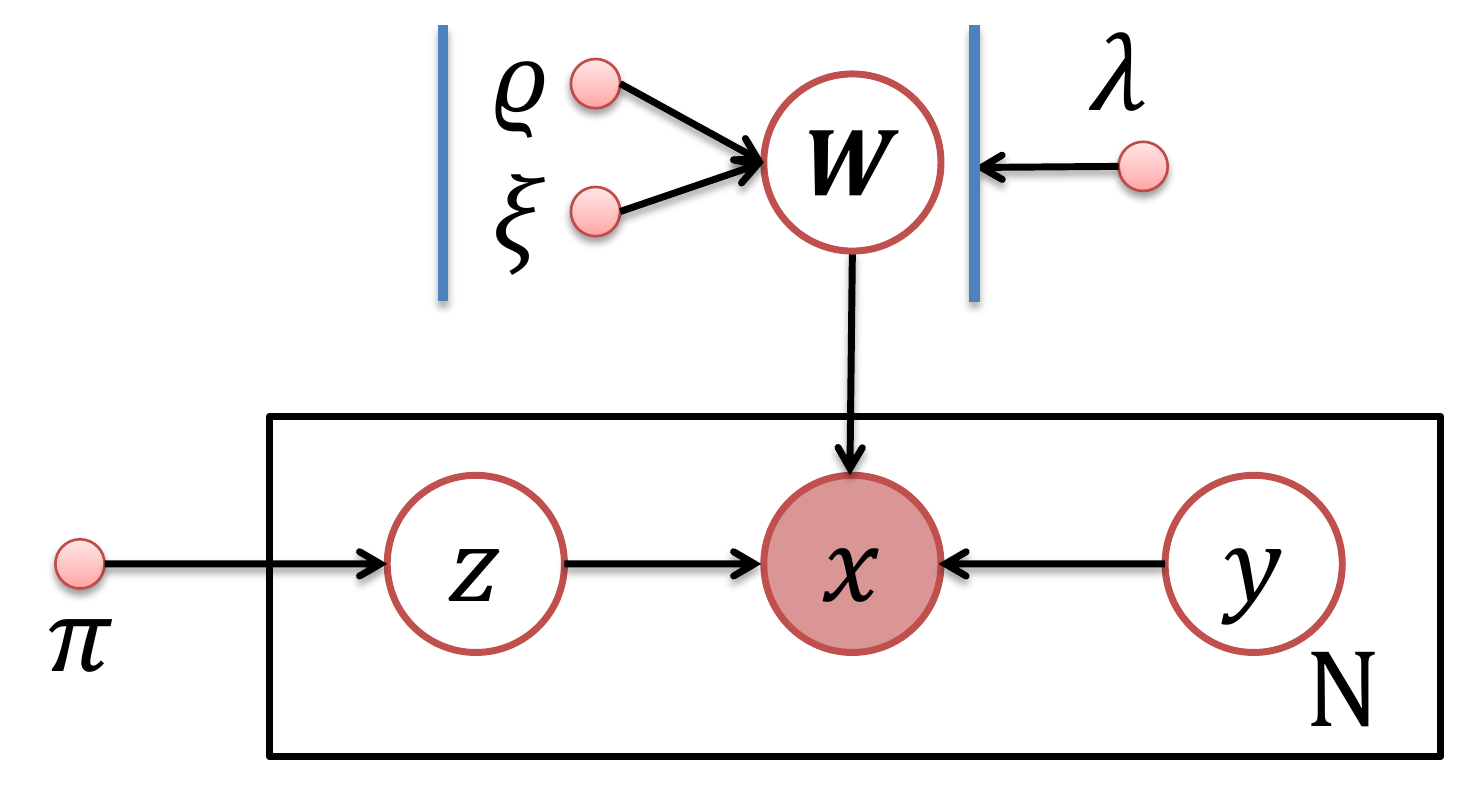}
  \caption{DEPCAM}
  \label{fig:defpcamm}
  \end{subfigure}
  \caption{Graphical representations of models: (a) SePCA-MM; (b) the proposed DEPCAM. The only difference between these two models is obviously the priors over $K$ mixture component parameters, i.e., $W$s. The traditional SePCA assigns independent isotropic Gaussian distributions parameterized with $\lambda$ and represented as a plate. Comparatively the proposed DEPCAM assigns a distribution over joint component parameters, i.e., $\mathbf{W} = \{W^1,\dots,W^K\}$. This joint distribution is a DPP parameterized with $\varrho,\xi$, and $\lambda$. It is represented as a double-struck.}
\end{figure*}

Formally, a layer of hidden discrete-valued variables $Z = \{z_n\}_{n=1}^N$ corresponding to observations $\{x_n\}_{n=1}^N$ is introduced to SePCA, as illustrated in the graphical representation shown in Figure \ref{fig:efpcamm}. Let $K$ be the number of mixing components. The value of $z_n\in \{1,\dots,K\}$ is the index of which mixing component each observation $x_n$ belongs to. {It obeys a multinomial distribution and parameterized by $\pi$}.
Comparing to the traditional SePCA, $K$ rather than $1$ transformation matrices $\mathbf{W}=\{W^1,\dots, W^k, \dots, W^K\}$ are introduced, and represented by a plate symbol in its graphical representation (Figure \ref{fig:defpcamm}).
The joint distribution over observable and hidden variables as well as transformation matrix variables is formulated as below.
\begin{align}
p(X,Y,Z,\mathbf{W}; &\pi,{\zeta}) \nonumber \\
&= p(X|Y,\mathbf{W},Z) p(Y) p(\mathbf{W};{\zeta}) p(Z;\pi).  \nonumber
\end{align}
As most of the probability distributions are the same with SePCA, we only briefly summarize the changes of SePCA-MM as follows.
The first term defines the likelihood of observations after observations are assigned to components. Take one observation $x_n$ as an example. When $z_n=k$, its likelihood has the same form with (\ref{eq:SePCALikelihood}) but substituting $W$ with a specified component $W^{k}$. The mixing concept is reflected by a proportional integration over all mixing components, and is formally given by
\begin{align}
p(x_n|y_n,\mathbf{W};\pi) = \sum_{z_n} p(z_n;\pi) p(x_n|y_n,{W}^{z_n}).  \nonumber
\end{align}
The likelihood over joint observations $X$ is obtained by multiplying over all likelihoods of individual observations since they are assumed to be independent given all hidden variables.  The second term assigns an isotropic unit Gaussian prior distribution to each hidden low-dimensional representation, which is the same with that in (\ref{eq:SePCA}). The third term gives a prior over $K$ mixing transformation matrices $\mathbf{W}$. It assumes that each mixing component is independent from each other, and that each PC within one mixing component is also independent to encode the orthogonality of a transformation matrix, and has the same prior distribution with (\ref{eq:singleWprior}). The fourth term presents a multinomial prior for each discrete-valued indicator variable, and the joint distribution over $Z$ is formulated as $p(Z;\pi) = \prod_{n=1}^N p(z_n;\pi)$ with $\sum_{k=1}^K \pi_k = 1$.

However, mixture models have two main limitations and cannot do well in some application scenarios. The first one is that the fitted mixing components may overlap with each other and cause ambiguity. This is unfavourable by real-world applications. For instance, when applying mixture models to clustering tasks, each component is used to model one cluster. The overlap between mixing components may lead to improper data separation, and ultimately result in ambiguous clustering results. In a similar way, this limitation is also not tolerated by `gaps' requiring applications, e.g. species delimitation \cite{yang2010bayesian}. Another limitation is that it may be fit with more components than it really needs to cover the whole observation space, which would easily lead to over-fitting. Under those circumstances, a diversity-encouraging prior encouraging repulsiveness amongst mixing components is in demand to mitigate this problem. In this paper, we focus on alleviating the above `overlapping' problem. We show how to define a diversity-encouraging prior for SePCA-MM as well as how to do model inference and parameter learning in next sections. Before that, we briefly introduce DPP utilized to build a diversity-encouraging prior in our method.

%

\subsection{DPP}

Determinantal point processes (DPP) are important statistical tools for diverse/repulsive relationship modelling.
It has been popular in machine learning area for years attributed to its tractable and efficient inferences \cite{Alex2012}.
Here we briefly review a DPP defined by $L$-ensembles \cite{Alex2012,borodin2005eynard} to establish a background for our study.

Given a ground data set $\mathcal{Y}$ (can be both continuous \cite{RajaHafiz20132} and discrete \cite{Alex2012}) and a real symmetric matrix $L$ indexed by the elements of $\mathcal{Y}$,
a point process $\mathcal{P}$ is called a determinantal point process if a random subset $\mathbf{Y}$ draws a sample $Y$ according to
\begin{align}
\mathcal{P}_{{L}}(\mathbf{Y}=Y) \propto \det({L}_{Y}),   \nonumber
\end{align}
where $Y \subseteq \mathcal{Y}$, and
${L}_Y \equiv [\mathcal{Y}_{ij}]_{i,j\in Y}$ denoting a submatrix of ${L}$, whose entries are indexed by the elements of $Y$. That $\det({L}_{\emptyset})=1$ is adopted from the convention.
The normalization of above equation is a constant once the matrix $L$ is fixed. Therefore, we simply ignore it.
More details and properties of DPPs for interested readers are referred to the references \cite{Christophe2014}\cite{JenniferGillenwater2012}\cite{HoughBen2006}.

\section{Proposed DEPCAM}
\label{sec:ourmodel}

\subsection{Motivation for Matrix-valued DPP}
There might exist overlap amongst the mixing components of traditional SePCA-MM, and imposing a diversity-encouraging prior over them is a natural strategy to alleviate this problem.
However, it does not exist a proper diversity kernel matrix $L$ defined over the mixing components of SePCA-MM in the current literature.
A probability measure based kernel can be considered applicable \cite{JamesZou2012} when combined with a prior distribution for each transformation matrix as employed in \cite{li2013simple}. Such a prior distribution employs an isotropic Gaussian distribution controlled by precision hyper-parameters, which enables the model automatically determine the effective number of PCs via an ARD scheme.
However, it is not ideal for our situation. As the means of the isotropic Gaussian priors are all fixed to $0$, it is ambiguous to infer group labels for samples around $0$.
In addition, a diversity kernel matrix constructed with a probability measure over such prior distributions is only relevant to the precision hyper-parameters, and has no relevant terms for transformation matrices $\mathbf{W}$ themselves. Consequently, this kind of kernel cannot provide an intuitive geometric explanation for diversity amongst these mixing components.
Therefore, a diversity kernel over transformation matrices, i.e., mixing components of SePCA-MM, serving as a fundamental block of a diversity-encouraging prior needs to be customized.

\subsection{Matrix-valued DPPs}

In this subsection, we construct a diversity-encouraging prior over the mixing components of PCA mixture models, parameterized with a set of matrices $\mathbf{W} = \{W^1, \dots, W^K\}$ with three steps - decomposing each transformation matrix into quality and similarity terms, formulating the quality and similarity terms, and integrating these two terms into the DPP framework to define the diversity-encouraging prior.

\subsubsection{Decomposition}
Each transformation matrix $W$ is decomposed into two parts: An orthonormal matrix $\Upsilon \in \mathcal{R}^{D\times d}$ whose columns representing elementary PCs and a diagonal matrix $\Phi \in \mathcal{R}^{d\times d}$ each of whose entries representing variances along each elementary PC. Formally,
\begin{align}
W=\Upsilon \Phi ,
\label{eq:decom}
\end{align}
with $\Upsilon^\top\Upsilon=I$ and
$\Phi=\mathrm{diag}({\Phi_1,\dots,\Phi_d})$.
To make the mathematical symbols clear, we apply the plain font for matrices (as in (\ref{eq:decom})) and the bold font for matrix sets, e.g., a $K$-size transformation matrix set $\mathbf{W}=\{W^1,\dots,W^K\}$, a $K$-size orthonormal matrix set $\mathbf{\Upsilon} = \{\Upsilon^1,\dots,\Upsilon^K\}$  and a $K$-size diagonal matrix set $\mathbf{\Phi}=\{\Phi^1,\dots, \Phi^K\}$ respectively.

Such a decomposition splits the orthogonality and the variance of a transformation matrix, and is motivated by three advantages. Firstly, the orthonormal part explicitly keeps the orthogonality amongst PCs within PCAs. In other words, the dimensions in low-dimensional space transformed by these PCs retain uncorrelated, which is a fundamental characteristic of PCAs for dimensionality reduction. Secondly, the variance part plays a role in summarizing variances along PCs. The higher the variance is, the more important the corresponding PC is. Therefore, this part provides a straightforward but effective way to automatically decide the effective number of PCs via retaining important PCs and neglecting insignificant ones. Thirdly, it provides a convenient way to define a diversity kernel $L$ by constructing a quality term and a similarity term separately, as presented in the subsequent sections.

\subsubsection{Formulation for quality and similarity terms}
Similar to \cite{Alex2012}, a quality-similarity decomposition is employed to construct a similarity kernel matrix, which is employed for the DPP-based diversity-encouraging prior. Such decomposition explicates the trade-off between element-wise quality and set-wise diversity.
These two terms in our case are respectively defined over variance and orthonormality matrices from above decomposition.

\textbf{Quality:}
We refer to the variance part $\Phi$ of a transformation matrix $W$ as its quality features.
These quality features build up the quality term of $W$. Formally, it is formulated as:
\begin{align}
\mathcal{Q}(W) = \exp (-\frac{1}{2}\xi  ||\Phi||_1).  \nonumber
\end{align}
Here $||\cdot||_1$ symbols the matrix $\ell_1$ norm. $\xi$ is a scale parameter to control two kinds of trade-offs - one between quality and diversity and the other one between diversity and likelihood.
$\ell_1$ regularization has been extensively used as sparsity constraints \cite{candes2008enhancing}. Here, we apply it to automatically determine dominating PCs of each mixing component. This strategy is inspired by ARD with zero-meaned Gaussian priors. Instead of choosing PCs with small variances, here, PCs with nearly zero values enforced by $\ell_1$ are neglected.

An $\ell_1$-norm, which is usually employed to encourage sparsity, is applied here to penalise the diagonal variance matrix $\Phi$.
It can be seen that high quality scores can only be obtained with a small $\ell_1$-norm value of the diagonal variance matrix, which equals to that many of the diagonal entries of the variance matrix are extremely small or exact $0$s. In other words, their corresponding PCs are insignificant and can be abandoned during the dimensionality reduction process. Therefore, this quality term encourages extremely low-dimensional hidden spaces. Furthermore, as mentioned above, this mechanism automatizes the process of model selection.

\textbf{Similarity:}
We refer to the orthonormal part $\Upsilon$ of a transformation matrix $W$ as its similarity features. The similarity function $\mathcal{S}$ over pairwise transformation matrices $\{W^1, W^2\}$ is constructed via their separable similarity features $\{\Upsilon^1, \Upsilon^2\}$, and is defined as
\begin{align}
\mathcal{S}(W^1, W^2) &= \exp(- \frac{1}{2} \varrho  \sum_{i,j=1}^d||\Upsilon^1_{\cdot i} - \Upsilon^2_{\cdot j}||^2_2), \label{eq:kernel} \\
s.t.~\Upsilon^{1T} \Upsilon^{1} &= I_d,   \nonumber \\ 
~ \Upsilon^{2T} \Upsilon^{2} &= I_d.  \nonumber 
\end{align}
Here $\varrho$ is a scale parameter, and
$||\cdot||_2$ symbols the vector $\ell_2$ norm, $I_d$ is a $d \times d$-sized identity matrix, and $d$ is the dimension of lower-dimensional spaces.
The two constraints guarantee the orthonormality of $\Upsilon$s.
%
\begin{figure*}[!ht]
\centering
\includegraphics[width=1\textwidth]{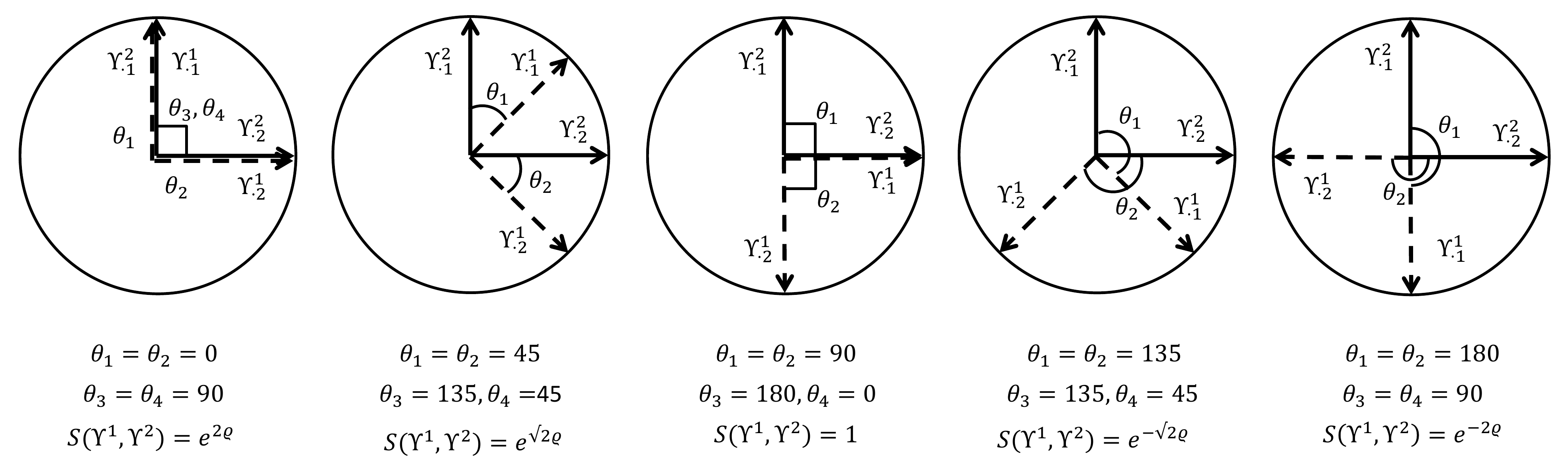}
\caption{Illustration of proposed similarity formulations with two orthonormal matrices including two PCs.
The  dashed pairwise perpendicular lines represent PCs of orthonormal matrix $\Upsilon^1$, while the solid pairwise lines represent PCs of $\Upsilon^2$.
$\theta_1$ and $\theta_2$ represent angles of direct-matching pairwise PCs as shown in all subfigures, while $\theta_3$ and $\theta_4$ represent angles of cross-matching pairwise PCs which are only demonstrated in the first subfigure and ignored by other subfigures to keep their symbols less crowded.
From left to right, the angles between direct-matching pairwise PCs of the two orthonormal matrices starts from $0^{\circ}$ and increases with $45^\circ$, while the similarities between these two matrices decreases from the largest value $e^{2\varrho}$ to the smallest value $e^{-2\varrho}$, which are calculated from (\ref{eq:similarity2}).
}
\label{fig:similarityIllustration}
\end{figure*}
Expanding the vector $\ell_2$ norm, the similarity function $\mathcal{S}$ is derived as:
\begin{align}
\mathcal{S}(W^1, W^2) &= \exp (- \frac{1}{2} \varrho  ( \sum_{i,j=1}^d \sum_{k=1}^D  (\Upsilon^1_{ki}-\Upsilon^2_{kj})^2 )) \nonumber
\\
&\propto \exp (\varrho  \sum_{i,j=1}^d \cos(\langle \Upsilon^{1}_{\cdot i} , \Upsilon^{2}_{\cdot j} \rangle)  ). \label{eq:similarity2}
\end{align}
Here $D$ is the dimension of the observation space, and $\langle \cdot,\cdot \rangle$ defines the angle between two vectors.
The only difference between (\ref{eq:similarity2}) and (\ref{eq:kernel}) is a constant term. This term is exactly the same for all entries of the similarity matrix, therefore does not make contributions to the diversity modelling and is ignored in the following modelling procedure.
Just like \cite{XieZhuXing2016}, this similarity is built on angles of pairwise vectors. The difference is that here it is a basic constructive element for DPP to further build diversity, while Xie et al. directly make use of them to formulate the set diversity. 



\textbf{Geometrical explanation for the similarity formulation:}
As it is easily seen from (\ref{eq:similarity2}), the similarity between two transformation matrices is positively related to the sum of the cosine values of the angles formed by pairwise normalized PCs. In other words, the similarity is negatively related to the angles between pairwise PCs. An example of such a similarity formulation is illustrated in Figure \ref{fig:similarityIllustration}. It demonstrates the similarity between two orthonormal matrices where each matrix consists of two PCs.
When the two matrices are the same, i.e., $\Upsilon^1=\Upsilon^2$, the angles for direct-matching pairwise PCs, i.e., $\Upsilon^1_{\cdot 1} \text{-} \Upsilon^2_{\cdot 1}$ and $\Upsilon^1_{\cdot 2} \text{-} \Upsilon^2_{\cdot 2}$, are $0^{\circ}$; and the angles for cross-matching pairwise PCs, i.e., $\Upsilon^1_{\cdot 1} \text{-}\Upsilon^2_{\cdot 2} $ and $\Upsilon^2_{\cdot 1} \text{-} \Upsilon^1_{\cdot 2}$, are $90^{\circ}$ due to the orthonormality of $\Upsilon$s. Calculating via (\ref{eq:similarity2}),
the similarity between them is $\exp(2\varrho)$, which is the largest. The similarities of two transformation matrices with other angles for direct-matching pairwise PCs, such as $45^{\circ}, 90^{\circ}, \text{~and~} 135^{\circ}$, have different similarity values, such as $e^{\sqrt{2} \varrho}, 1$, and $e^{-\sqrt{2}\varrho}$ respectively.
Particularly, when each column $\Upsilon^1_{\cdot j}$ of a matrix $\Upsilon^1$ is totally different with direct-matching column $\Upsilon^2_{\cdot j}$ of matrix $\Upsilon^2$, the angle formed by them is $180^{\circ}$. The corresponding cosine value is $-1$, and the resulting similarity between the two matrices is $e^{ -2\varrho }$, which is the smallest.

\subsubsection{Diversity-encouraging prior}		
Each entry of an ensemble kernel $L$, defined over the transformation matrix set, represents the similarity of its indexed two transformation matrices. The entry value is defined on its two relevant matrices by multiplying their quality terms with similarity term \cite{Alex2012}. Formally,
\begin{align}
\mathcal{L} (W^1, W^2) = \mathcal{Q}(W^1) \mathcal{S}(W^1,W^2) \mathcal{Q}(W^2). \nonumber
\end{align}
A DPP distribution over transformation matrices is defined via a determinant operator over the $L$-ensemble kernel ${L}$.
For example, the DPP probability over a two-element transformation matrix set $\{W^1, W^2\}$ is computed as
\begin{align}
\mathcal{P}_{{L}}(\{W^1,&W^2\})  \propto  \det (\mathcal{L} (\{W^1, W^2\})) \label{eq:final1} \\
&\propto \mathcal{Q}^2(W^1) \mathcal{Q}^2(W^2) \det(\mathcal{S}(\{W^1,W^2\})) \label{eq:final2}
\\
&=\exp (- \xi (||\Phi^1||_1 + ||\Phi^2||_1) )
\nonumber \\
&
\times \left\{1 - \exp (2\varrho  \sum_{i,j=1}^d \cos(\langle\Upsilon^{1}_{\cdot i} , \Upsilon^{2}_{\cdot j}\rangle ) )
\right\},
\label{eq:diverseProbability2}
\end{align}
where $\mathcal{L} (\{W^1, W^2\})$ and $\mathcal{S}(\{W^1,W^2\})$ represent the L-ensemble kernel matrix and similarity matrix indexed by $W^1$ and $W^2$ respectively.
{In \eqref{eq:diverseProbability2}, the first term combines the quality measure of two elements, while the second term is the determinant of a $2\times 2$ symmetric matrix whose diagonals are $1$ and anti-diagonal elements are from \eqref{eq:similarity2}.}
Note that the normalization term is ignored here, since it is a constant for various subsets once the parameters $\xi$ and $\varrho$ are given. {Extending this to more than two elements set $\mathbf{W}=\{W^1,W^2,\dots\}$ is straightforward. Formally,}
{
\begin{align}
\mathcal{P}_{L}(\mathbf{W};\varrho,\xi) \propto \det(\mathcal{L}(\mathbf{W})). \notag
\end{align}
Whenever no confusion arises, $\varrho$ and $\xi$ are omitted to simplify notations. }

We draw two characteristics of the above definition.
{First}, the computations for the quality term and similarity term are independent, due to one nice property of the determinant operator that $\det(AB) =\det(A)\det(B)$ for square matrices $A,B$ as applied from (\ref{eq:final1}) to (\ref{eq:final2}).
{Second}, from (\ref{eq:diverseProbability2}), we can see that the diverse probability over a transformation matrix set increases along with the increment of the quality of each element as well as with the decrement of the joint similarities amongst them.

In summary, this DPP distribution prefers subsets of transformation matrices that
its mixing components are diverse with each other as well as
each mixing component is of small number of PCs.

\subsection{DEPCAM}
Finally, by combining the established diversity-encouraging prior into the traditional SePCA-MM, its diversified version, i.e., DEPCAM, is established. Its graphical representation is shown in Figure \ref{fig:defpcamm}. Two bold vertical lines drawing over $\mathbf{W}$ represent the diversity-encouraging prior introduced to the joint mixing components of SePCA-MM.
Formally,
the joint distribution over both observable and hidden variables is formulated as:
\allowdisplaybreaks
\begin{align}
 p(X,Y,Z,\mathbf{W}; \pi,& \lambda, \varrho, \xi)  \nonumber \\
 \propto  \prod_{n=1}^N p(z_n;\pi) & p(y_n) p(x_n|y_n, W^{z_n}) (\mathcal{P}_{{L}}(\mathbf{W};\varrho,\xi))^{^{\lambda}}\label{eq:overallModel1}\\
s.t.\qquad &  W^k =  \Upsilon^k \Phi^k, \quad k=1,\dots,K , \nonumber
\\
&\Upsilon^{k\top} \Upsilon^k = I_d,  \label{eq:orthonormalUpsilon}\\
& \Phi^k = \mathrm{diag}({\Phi^k_1, \dots, \Phi^k_d}) , \nonumber
\\
&\sum_{k=1}^K \pi_k = 1.   \nonumber
\end{align}
Here
$\mathbf{W}$ is a $K$-sized transformation matrix set. Each element of it, i.e., $W^k$, is decomposed into an orthonormal matrix $\Upsilon^k$ and a diagonal matrix $\Phi^k$, referred to as its diverse features and quality features respectively. $K$ is the number of mixture components.
The constraint for $\pi$ is to satisfy a discrete categorical distribution.
The diversity-encouraging prior $\mathcal{P}_L(\mathbf{W};\varrho,\xi)$ is changed to its power of $\lambda$ of itself. This new parameter $\lambda$ plays a trade-off role in balancing the diversity-encouraging prior and model fitness as explained below.

The parameters and their functionalities are summarized as follows.
$\xi$ and $\varrho$ are introduced in the diversity-encouraging prior (\ref{eq:diverseProbability2}), wherein $\xi$ is for the quality measure of individual transformation matrix and $\varrho$ is for the similarity measure of pairwise transformation matrices.
When combined into the diversity-encouraging prior, they can be adjusted to balance quality and similarity within a transformation matrix set.
Also these two parameters play a trade-off role when they are as part of the whole model in (\ref{eq:overallModel1}). Specifically, they are utilized to make a balance between the diversity of $\mathbf{W}$ and the model fitness. To make this trade-off role clear, we add a new trade-off parameter $\lambda$, as shown in (\ref{eq:overallModel1}).
It makes no substantial change to the original model, but makes the efficacy demonstration of the proposed diversity-encouraging prior easy in the experimental section.

Note that the parameters $\xi$ and $\varrho$ are fixed in the model for parameter learning and inference, but can be empirically chosen to be adapted to real-world datasets.
Actually, this is a simplification for our model to avoid intractable learning for these two parameters, since they also exist in the normalization term of the diversity-encouraging prior.
One drawback of this simplification is that it limits us to assigning point estimations for $\mathbf{W}$ rather than doing Bayesian inference. Their learning and inference procedures are shown in the next section.

\section{Learning and inference}
\label{sec:learning}
We address parameter learning and inference of the proposed model within an expectation-maximization (EM) framework. Specifically, the parameter $\pi$ is learned in M-step via maximum likelihood estimation (MLE). In E-step, the posterior distribution of hidden variables $Z$ is inferred, while the point estimations for hidden variables $\mathbf{W}$ and $Y$ are approximately estimated via maximizing their posterior distributions.

\subsection{Learning Parameters: M-step}
\subsubsection{Learning $\pi$} The objective of log-likelihood maximization in terms of $\pi$ is
\begin{align}
\max_{\pi} \log p(X;\pi) 
&= \sum_{n=1}^N \log \sum_{z_n} p(x_n|z_n) p(z_n; \pi),
\label{eq:normalizationConPi0} \\
s.t.   \sum_k \pi_k &= 1. \label{eq:normalizationConPi}
\end{align}
where $p(x_n|z_n)$ is the $z_n$-conditional likelihood, and $p(z_n;
\pi)$ is the prior distribution of $z_n$. Directly applying Lagrange multiplier method is intractable due to the integration over $z_n$ inside the $\log$ operator. Therefore, we approximate the above log-likelihood with its lower bound function obtained by employing Jensen's inequality, which is
\begin{align}
\mathcal{L}_{q_Z}(\pi) =& \sum_{n=1}^N \sum_{z_n} q(z_n) \log p(x_n|z_n)  \label{eq:lowerbound} \\
& + \sum_{n=1}^N \sum_{z_n} q(z_n) \log p(z_n;\pi) + \mathcal{H}_{q_{zn}}, \nonumber
\end{align}
where $q(z_n)$ is a distribution over $z_n$ and $H_{q_{z_n}}$ is its entropy. The above lower bound is equal to in (\ref{eq:normalizationConPi0}) when $q(z_n)$ is of the posterior distribution over $z_n$.
Specifically, in above equation, the first term, expressing an empirical expectation over the conditional log-likelihood in terms of $q(z_n)$, and the third term are independent of the parameters $\pi$. Therefore, they could be simply ignored when learning $\pi$. Only the second term, which is an empirical expectation over log-$z_n$ prior in terms of $q(z_n)$, is used to learn $\pi$ subjecting to the constraint (\ref{eq:normalizationConPi}). Its closed-form solution is obtained by employing the Lagrange multiplier method and is:
\begin{align}
\pi_k = \frac{\sum_n q(z_n = k)}{\sum_k \sum_n q(z_n=k)} = \frac{N_k}{N}.
\label{eq:updatepi}
\end{align}
Here $N$ is the number of all samples, and $N_k$ is the value summarizing over probabilities of all samples in group $k$.
It is easily seen that $\pi_k$ is updated via the ratio between the cardinalities of group $k$ and the whole dataset.

\subsection{Inference: E-step}
\subsubsection{Posterior inference on $Z$}
The posterior distributions over $Z$ are computed through the Bayes' rule, namely,
\begin{align}
q(Z) = p(Z|X;\pi) = \frac{p(Z;\pi) p(X|Z)}{\sum_{Z}p(Z;\pi)p(X|Z)}.
\label{eq:posteriorZ}
\end{align}
Here $p(Z;\pi)$ and $p(X|Z)$ are quite the same with $p(z_n;\pi)$ and $p(x_n|z_n;\pi)$ in (\ref{eq:lowerbound}), but incorporates all samples rather than only one.
This marginal posterior distribution can be computed by integrating over two other hidden variable sets $Y$ and $\mathbf{W}$ by exploiting their known joint distribution. Formally
\begin{align}
p(X|Z)
&= \int_{Y}\int_{\mathbf{W}} p(Y) {\mathcal{P}_{{L}}(\mathbf{W})}p(X|Y, \mathbf{W},Z).
\label{eq:YWintegration}
\end{align}
However, either of the two integrations has a closed form.
As for the integration along $Y$, there is a generalized exponential family term in the $\{Y,\mathbf{W},Z\}$-conditional likelihood $p(X|Y,\mathbf{W},{Z})$, which is not conjugated with its prior distribution.
Similarly, for the integration along $\mathbf{W}$, its DPP-based prior distribution has no conjugate property with exponential family terms.
Intuitively, these two integrations can be approximated via a Monte Carlo method.
It firstly draws two sets of samples from $Y$ and $\mathbf{W}$'s prior distributions, respectively, and secondly substitutes the samples for variables to compute the conditional likelihood, and finally replaces the integration with a tractable average operator over those samples' likelihoods.
However, although the prior distribution of $p(Y)$ can be easily sampled,
the joint transformation matrix set $\mathbf{W}$ cannot be easily obtained from its DPP-based prior distribution.

Therefore, we again approximate the above $\{Y,\mathbf{W}\}$-marginalized conditional likelihood with its lower bound in $\log$ space.
With Jensen's inequality,
one
lower bound to (\ref{eq:YWintegration}) is obtained as
\begin{align}
\mathcal{L}_{q}(\mathbf{W},  &Y) =
\left\{ \mathcal{H}_{q_{\mathbf{W}}} + \mathcal{H}_{q_Y} \right. \nonumber\\
&\left. +
 \int_{\mathbf{W}} \int_{Y}
q(\mathbf{W}) q(Y) \log p(X,Y,\mathbf{W}|Z)  \right\} \label{eq:lowerboundWY}.
\end{align}
Here $q(\mathbf{W})$ and $q(Y)$ are $\mathbf{W}$ and $Y$'s posterior distributions respectively.
The first two terms $\mathcal{H}_{q_{\mathbf{W}}}$ and $\mathcal{H}_{q_Y}$ are entropies of posterior distributions of $\mathbf{W}$ and $Y$ respectively.
The third term computes the expectation over $\log$ $Z$-conditional joint distribution in terms of the posterior distributions of $\mathbf{W}$ and $Y$.
Instead of integrating along $\mathbf{W}$ and $Y$'s posterior distributions, which is intractable, we solve it by an efficient approximation, namely substituting the posterior distributions with their posterior mode estimation. Formally
\begin{align}
& \mathcal{L}_{q}(\mathbf{W},Y)  \approx
\log p(X,Y^{\mathrm{MP}},\mathbf{W}^{\mathrm{MP}}|Z) + \mathrm{const},
\end{align}
where the $\mathrm{const}$ summarizes over entropy terms.
Here, we assume that the variational posterior distributions of $Y$ and $\mathbf{W}$ are peaked at their most-probable values, namely $q(Y) \approx \delta(Y^{MP})$, and $q(\mathbf{W}) \approx \delta(\mathbf{W}^{MP})$ with $\delta$ the Dirac delta function. Such point estimations might not be representative for the true distributions. However, due to its attractive computational efficiency, we still favours it as our first choice. The experimental results confirm its effectiveness.


\begin{algorithm}[t]
 \caption{Parameter Learning and Inference}
 \KwData{Observations $X$, fixed parameters $\xi,\varrho$, stopping criterion $\epsilon$.}
 \KwResult{$\mathbf{W}^{\mathrm{MP}}$, $Y^{\mathrm{MP}}, \mathcal{L}^{\mathrm{new}}$, q(Z), $\pi$.}
 Randomly initialization $\pi$, $\mathbf{\Phi}^{\mathrm{MP}}$, $\mathbf{\Upsilon}^{\mathrm{MP}}$, $Y^{\mathrm{MP}}$;\; \\
 ${W}^{k,\mathrm{MP}}= {\Upsilon}^{k,\mathrm{MP}} \Phi^{k,\mathrm{MP}}$, $k=1,\dots,K$; \\
  $q(Z)$ $\leftarrow$ (\ref{eq:posteriorZ}) and $p(X,Y^{\mathrm{MP}},\mathbf{W}^{\mathrm{MP}}|Z)$; \\
 $\mathcal{L}^{\mathrm{new}} =\mathbb{E}_{ q(Z)}  (\log p(Z;\pi) + \log p(X,Y^{\mathrm{MP}}, \mathbf{W}^{\mathrm{MP}}|Z))$;  \; \\
\Repeat{ $|\mathcal{L}^{\mathrm{new}} - \mathcal{L}^{\mathrm{old}}|<\epsilon$ }
 {
   	$ \mathcal{L}^{\mathrm{old}} = \mathcal{L}^{\mathrm{new}}$ ; \\
   	M-step: \; \\
   	$\pi \leftarrow$ (\ref{eq:updatepi}) and $q(Z)$; \; \\
   	E-step:\; \\
	$\mathcal{L}^{\mathrm{new}}_0 = \mathbb{E}_{q(Z)}  (\log p(Z;\pi) + \log p(X,Y^{\mathrm{MP}}, \mathbf{W}^{\mathrm{MP}}|Z)); $  \\
   	\Repeat{$|\mathcal{L}^{\mathrm{new}}_0 - \mathcal{L}^{\mathrm{old}}_0|<\epsilon$}
   	{
		$\mathcal{L}^{\mathrm{old}}_0 = \mathcal{L}^{\mathrm{new}}_0$; \\
       		$q(Z) \leftarrow $ (\ref{eq:posteriorZ}) and $p(X,Y^{\mathrm{MP}},\mathbf{W}^{\mathrm{MP}}|Z)$ ; \\
       		Alternatively update $\Upsilon^k$ and $\Phi^k$ with (\ref{eq:derivationUpsilon}) and (\ref{eq:derivationPhi});\; \\
		${W}^{k,\mathrm{MP}}= {\Upsilon}^{k,\mathrm{MP}} \Phi^{k,\mathrm{MP}}$, $k=1,\dots,K$; \\
		$Y^{\mathrm{MP}} \leftarrow$ (\ref{eq:derivationY}); \\
  		$\mathcal{L}^{\mathrm{new}}_0 = \mathbb{E}_{q(Z)} (\log p(Z;\pi) ~+~  \log p(X,Y^{\mathrm{MP}}, \mathbf{W}^{\mathrm{MP}}|Z)); $  \\

  	}
  $\mathcal{L}^{\mathrm{new}} =\mathbb{E}_{q(Z)} (\log p(Z;\pi) ~+~ \log p(X,Y^{\mathrm{MP}}, \mathbf{W}^{\mathrm{MP}}|Z))$;
}
\label{alg:paraLandinfer}
\end{algorithm}

\subsubsection{Variational inference scheme} ~\\
 \vspace{0.2em}\noindent
\textbf{Posterior mode estimation for $\mathbf{W}$:}
The posterior distribution over $\mathbf{W}$ is formulated with its prior distribution and likelihood via Bayes' rule, namely,
\begin{align}
q(\mathbf{W}) = p(\mathbf{W}|X) &\propto p_{{L}}(\mathbf{W}) p(X|\mathbf{W}). \nonumber
\end{align}
Given $Y^{MP}$, the objective for mode estimation of posterior distribution of $\mathbf{W}$ in $\log$-space is
\begin{align}
\max_{\mathbf{W}} \mathcal{L}_q(\mathbf{W}) =
\log p_{{L}}(\mathbf{W})
+ \sum_Z q(Z) \log p(X,Y^{\mathrm{MP}}, Z|\mathbf{W}), \label{eq:LqW}
\end{align}
where the first term is related to $\mathbf{W}$'s diversity-encouraging prior distribution, and the second term is an expectation over log-likelihood of $\mathbf{W}$ in terms of $q(Z)$.

We apply a coordinate ascend method to iteratively search the optimal solution for the transformation matrix set $\{W^k\}_{k=1}^K$.
Within each coordinate,
since each ${W}^k$ is decomposed into two components, i.e., orthonormal matrix $\Upsilon^k$ and diagonal matrix $\Phi^k$, we alternatively update them.

\begin{figure*}[!ht]
\centering
\begin{subfigure}[b]{0.48\textwidth}
\includegraphics[width=\textwidth]{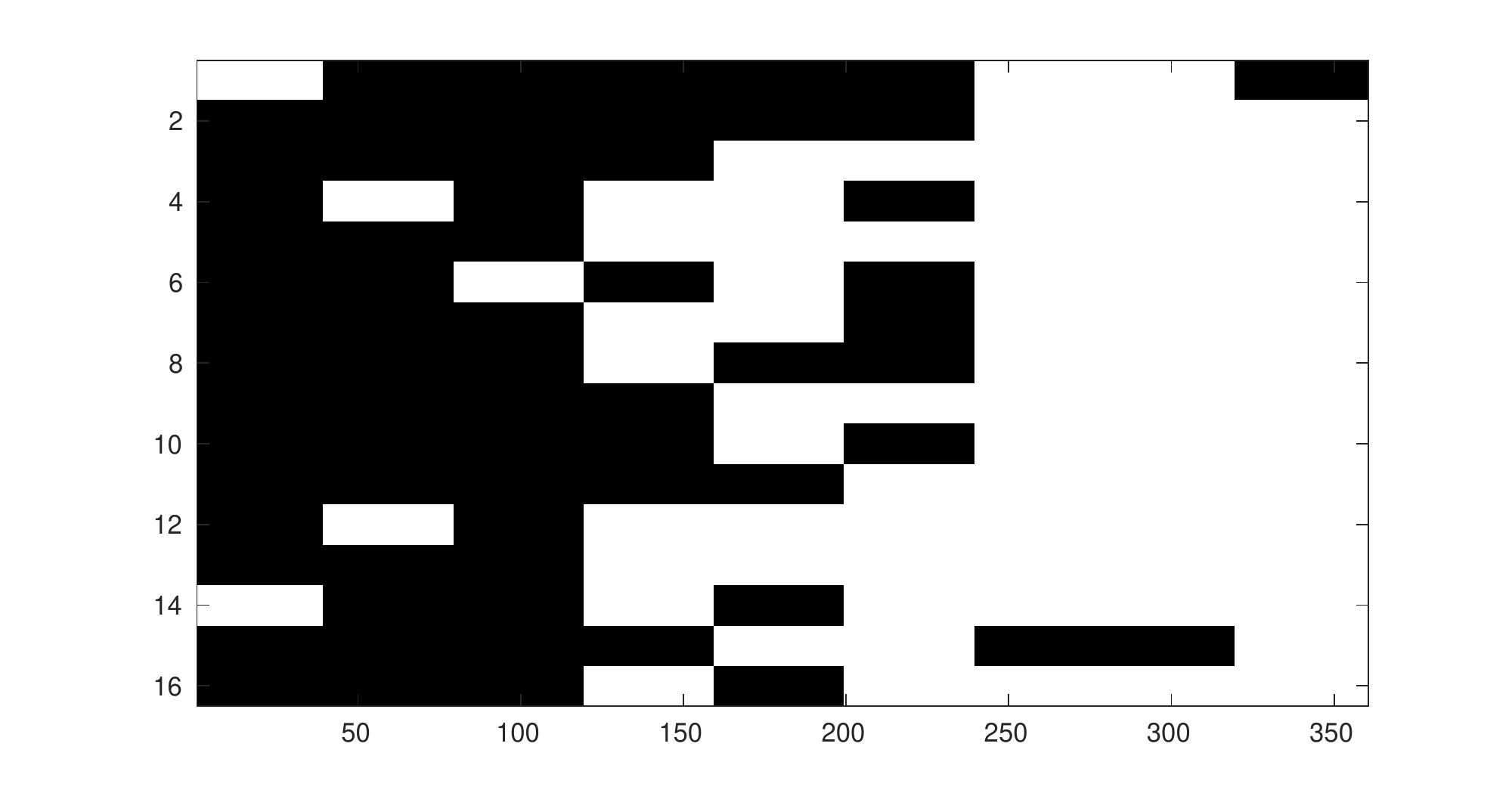}
\caption{True data}
\label{fig:trueD}
\end{subfigure}
\begin{subfigure}[b]{0.48\textwidth}
\includegraphics[width=\textwidth]{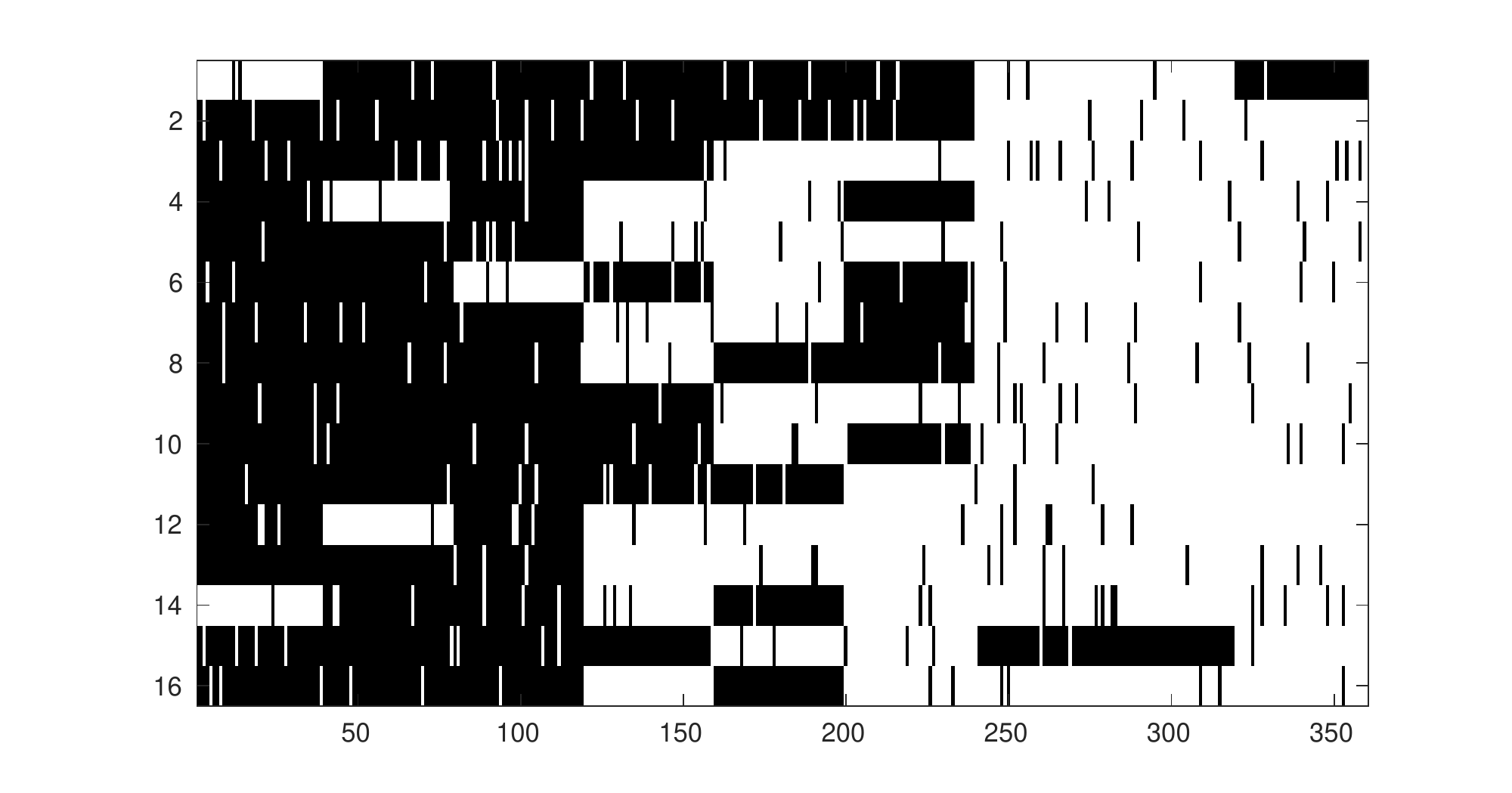}
\caption{Noisy data}
\label{fig:noisyD}
\end{subfigure}
\caption{Synthetic data with black indicating $0$ and white $1$. }
\end{figure*}

 \vspace{0.2em}\noindent \textbf{For} $\mathbf{\Upsilon}$\textbf{:}
Due to the orthonormal constraints (\ref{eq:orthonormalUpsilon}), we search its optimal along the Grassmann manifold.
The derivation in Euclidean space is
\begin{align}
&\frac{\partial \mathcal{L}_q(\mathbf{W})}{\partial \Upsilon^{k}_{ij}} = 
 \lambda \mathrm{trace}(
{L}^{-1}(\mathbf{W}) {\mathcal{Q}(\mathbf{W}) \frac{
\partial \mathcal{S}(\mathbf{W})
}{\partial \Upsilon^k_{ij}}
\mathcal{Q}(\mathbf{W})}) +  \nonumber \\  
& (q(Z=k))^\top \mathrm{diag}\left(g'^\top({\Upsilon}^k {\Phi}^k Y^{\mathrm{MP}} ) \cdot
\left( \frac{\partial {\Upsilon}^k} {\partial \Upsilon^k_{ij}}  {\Phi}^k Y^{\mathrm{MP}} \right) \right) \nonumber \\
&+
\mathrm{trace}(X^\top \frac{\partial {\Upsilon}^k}{\partial \Upsilon^k_{ij}} {\Phi}^k Y^{\mathrm{MP}} \mathrm{diag}(q(Z=k))),  
\label{eq:derivationUpsilon}
\end{align}
with each entry of $\frac{\partial \mathcal{S}({W})}{\partial \Upsilon^k_{ij}} $ from
\begin{align}
&\frac{\partial \mathcal{S}({W}^{k}, {W}^{k'})}{\partial \Upsilon^k_{ij}} =  
\mathcal{S}({W}^k,{W}^{k'}) \cdot \varrho \cdot \sum_m (\Upsilon^{k'}_{im} - \Upsilon^k_{ij}),
\nonumber
\end{align}
where $m$ traverses all columns of matrix $\Upsilon^{k'}$.
Let $G(\Upsilon^k) = \frac{\partial -\mathcal{L}_q(\mathbf{W})}{\partial \Upsilon^k}$,
the derivation on the Grassmann manifold is defined as:
\begin{align}
GG(\Upsilon^k) = G(\Upsilon^k) - \Upsilon^{k}\Upsilon^{k\top}G(\Upsilon^k).
\nonumber
\end{align}
At point $\Upsilon^{k}$ with direction $GG(\Upsilon^{k})$, the corresponding geodesic equation is
\begin{align}
\Upsilon^k(t)  = \Upsilon^k V \cos(\Sigma t) V^\top + U \sin (\Sigma t) V^\top,
\label{eq:geodesic}
\end{align}
where matrices $U$, $\Sigma$, and $V$ are from the compact SVD of $GG(\Upsilon^k)$, namely, $GG(\Upsilon^k) = U \Sigma V^\top$.
The optimal searching is along the geodesic defined by (\ref{eq:geodesic}).
This is actually a one-dimensional searching with respect to variable $t$, namely,
$\min_t -\mathcal{L}_q(\Upsilon^k(t) \Phi^k)$.
Suppose it reaches the minimum value at $t'$, then the corresponding point on the Grassmann manifold is $\Upsilon^{k'} = \Upsilon^k(t')$.

 \vspace{0.2em}\noindent\textbf{For $\mathbf{\Phi}$:}
We iteratively update $\{\Phi^1, \dots, \Phi^k,\dots,\Phi^K\}$ with gradient ascend method. The derivative of the objective (\ref{eq:LqW}) with respect to $\Phi^k_{i}$ is listed as below.
\\
\begin{align}
&\frac{\partial \mathcal{L}_q(\mathbf{W})}{\partial \Phi^{k}_{i}}
    = \mathrm{trace} ({L}^{-1}(\mathbf{W})
       \frac{\partial \mathcal{Q}(\mathbf{W})\mathcal{S}(\mathbf{W}) \mathcal{Q}(\mathbf{W})}
              {\partial \Phi_{i}^k})
       \nonumber \\ 
&
    + \mathrm{trace}(X^\top{\Upsilon}^k \frac{\partial {\Phi}^k}{\partial \Phi^k_{i}}Y^{\mathrm{MP}}
        \mathrm{diag}(q(Z=k)))  \nonumber \\ 
&
   +(q(Z=k))^\top
      \mathrm{diag}\left(g'^\top(\Upsilon^k \Phi^k Y^{\mathrm{MP}})
           \cdot  \Upsilon^k \frac{\partial \Phi^k}{\partial \Phi^k_{i}} Y^{\mathrm{MP}}\right).
 \label{eq:derivationPhi}
\end{align}
For term $
\frac{ \partial \mathcal{Q}({W}^k) \mathcal{S}(\{{W}^k, {W}^{k'} \}) \mathcal{Q}({W}^{k'}) }{\partial \Phi^k_{i}}
$, when $k=k'$, it is
\begin{align}
\frac{\partial \exp (- \xi  (||\Phi^k||_1))}{\partial \Phi^k_{i}} = 
\exp (- \xi  (||\Phi^k||_1)) \cdot -\xi \cdot \frac{\partial |\Phi^k_{i}|}{\partial \Phi^k_{i}}; \nonumber
\end{align}
otherwise, it is computed as
\begin{align}
\mathcal{S}(\{{W}^k, {W}^{k'} \}) \mathcal{Q}({W}^{k'})
\exp (- \frac{1}{2} \xi (||\Phi^k||_1)) \cdot -\frac{1}{2} \xi \cdot
\frac{\partial |\Phi^k_{i}|}{\partial \Phi^k_{i}}. \nonumber
\end{align}
Here $|\cdot|$ denotes absolute operator. As only the diagonal elements of $\Phi^k$ are non-zero,  $||\Phi^k||_1 = \sum_{i=1}^d|\Phi^k_{i}|$. It is easily seen that the derivative with respect to $\Phi^k_{i}$ in (\ref{eq:derivationPhi}) involves all other diagonal elements, i.e., $\{\Phi^k_{j}\}_{j\neq i}$. Therefore, we iteratively update the diagonal elements of $\Phi^k$  one by one.
Because the objective is nondifferientiable at $0$, a sub-gradient method is applied.
That the term $\frac{\partial |\Phi^k_{i}|}{\partial \Phi^k_{i}}$ is set to $-1$ or $1$ or $0$ depends on which value increases the objective most.

{Note that $\pi$ is initialized by a uniform-distributed sample, $\{\Phi^k\}_{k=1}^K$ by random diagonal matrices, and $\{\Upsilon^k\}_{k=1}^{K}$ by orthonormal matrices whose entries were first randomly generated and then processed with a Gram-Schmidt process.
These initializations are the same for both the standard SePCA-MM and the proposed diversified version for fair comparisons.}

\noindent \textbf{Posterior mode estimation for $Y$:}\\
Similarly to $\mathbf{W}$, the posterior distribution on $Y$ is computed as
\begin{align}
q(Y) \propto p(Y) p(X|Y).
\nonumber
\end{align}
Given $\mathbf{W}^{\mathrm{MP}}$, the objective for mode estimation of posterior distribution on $Y$ in log-space is
\begin{align}
\max_{Y} \mathcal{L}_q(Y) =  \log p(Y) +  \sum_Z q(Z) \log p(X,\mathbf{W}^{\mathrm{MP}},Z|Y).
\label{eq:objY}
\end{align}
To achieve the optimal of each independent $y_n$, the gradient ascend method is applied and its gradient is computed as below.
\begin{align}
\frac{\mathcal{L}_q(y_n)}{\partial y_n} =
&-y_n + \sum_{z_n} q(z_n)
 \left(   (x^\top_nW^{\mathrm{MP},z_n})^\top \right. \label{eq:derivationY}\\
 &\left.+  (g'^\top(W^{\mathrm{MP},z_n}y_n) \cdot W^{\mathrm{MP},z_n})^\top     \right).
\nonumber
\end{align}

The whole procedure of parameter learning and inference for the proposed method under a variational EM framework is summarized in Algorithm \ref{alg:paraLandinfer}.

\subsection{Prediction}
To do clustering over unseen test data, for example $x^*$, with the point estimation of transformation matrices $\mathbf{W}^{MP}$ from training dataset and learned parameters $\pi, \xi, \varrho$, we develop a predicting model as below,
\begin{align}
p(z^*|x^*, \mathbf{W}^{MP}; \pi, \xi,\varrho) = p(z^*|\pi)p(x^*|\mathbf{W}^{MP,z^*};\xi,\varrho).
\label{eq:predictionZ}
\end{align}
Here there is no closed-form for the second multiplier due to the existence of hidden variable $y^*$.
To make it computable,
just like in the training procedure, a mode point estimation, which is used to represent the posterior distribution of the low-dimensional representation of testing data $y^{*MP}$, is obtained via
\begin{align}
y^{*MP} = \max_{y^*} \log p(y^*|\mathbf{W}^{MP},x^*;\xi,\varrho,\pi),
\end{align}
where it can be solved as \eqref{eq:objY} does.
Finally, the second multiplier is approximated by $p(x^*|\mathbf{W}^{MP,z^*},y^{*MP};\xi,\varrho)$.

\begin{figure*}[!t]
\centering
\begin{subfigure}[b]{0.32\textwidth}
\includegraphics[width=\textwidth]{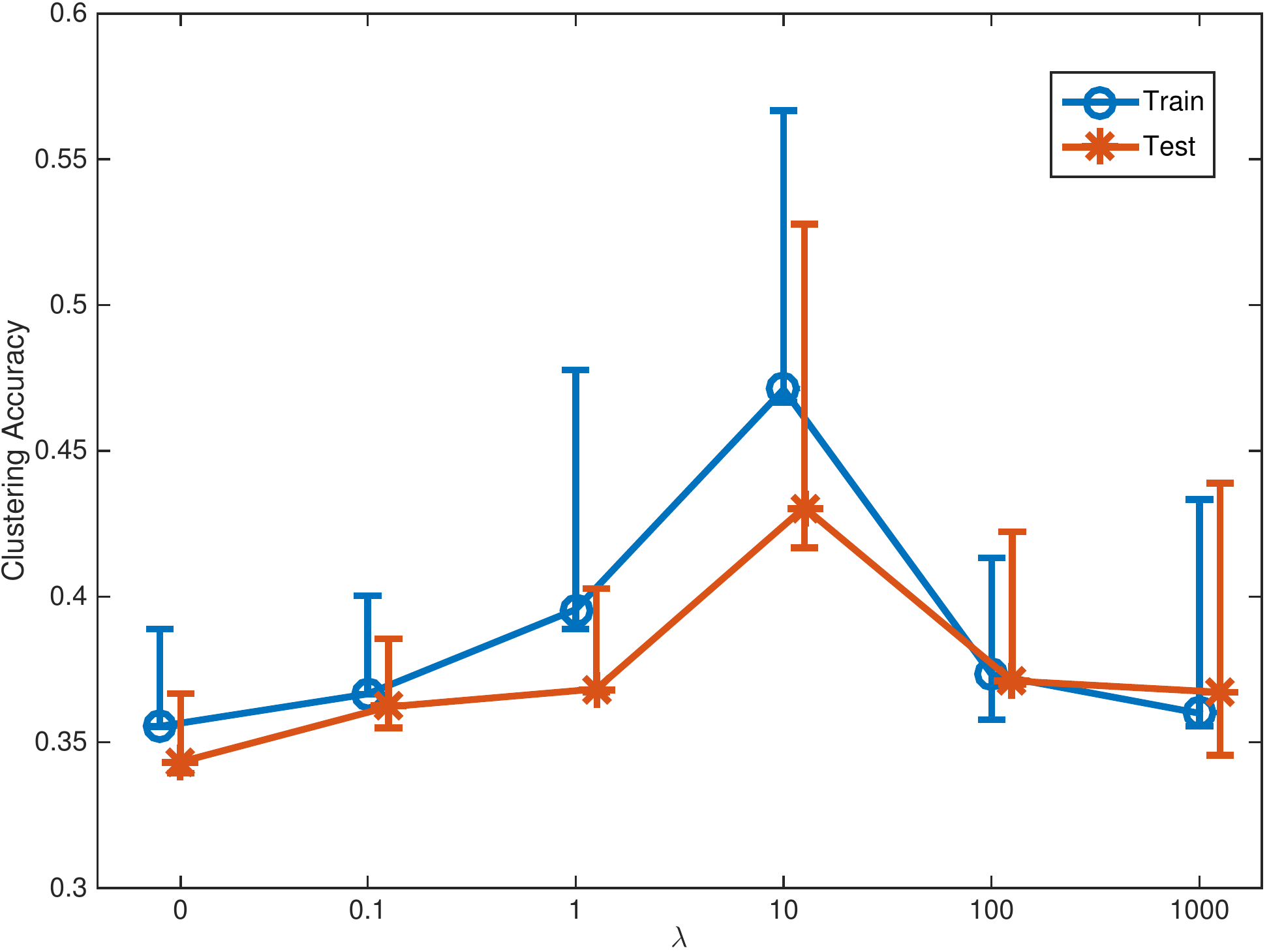}
\caption{Synthetic}
\label{fig:lamError}
\end{subfigure}
\begin{subfigure}[b]{0.32\textwidth}
\includegraphics[width=\textwidth]{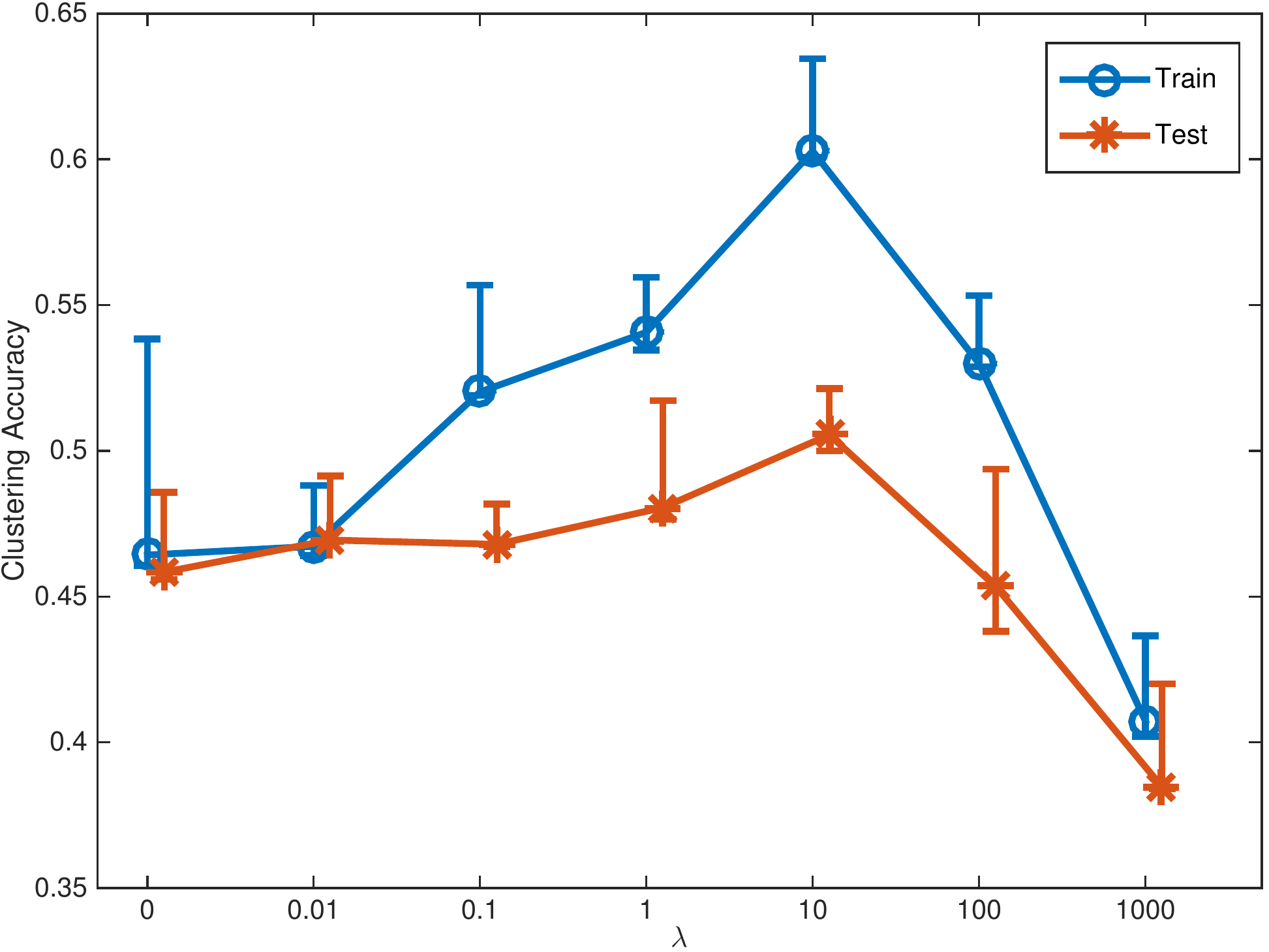}
\caption{USPS}
\label{fig:lambdaUSPS}
\end{subfigure}
\begin{subfigure}[b]{0.32\textwidth}
\includegraphics[width=\textwidth]{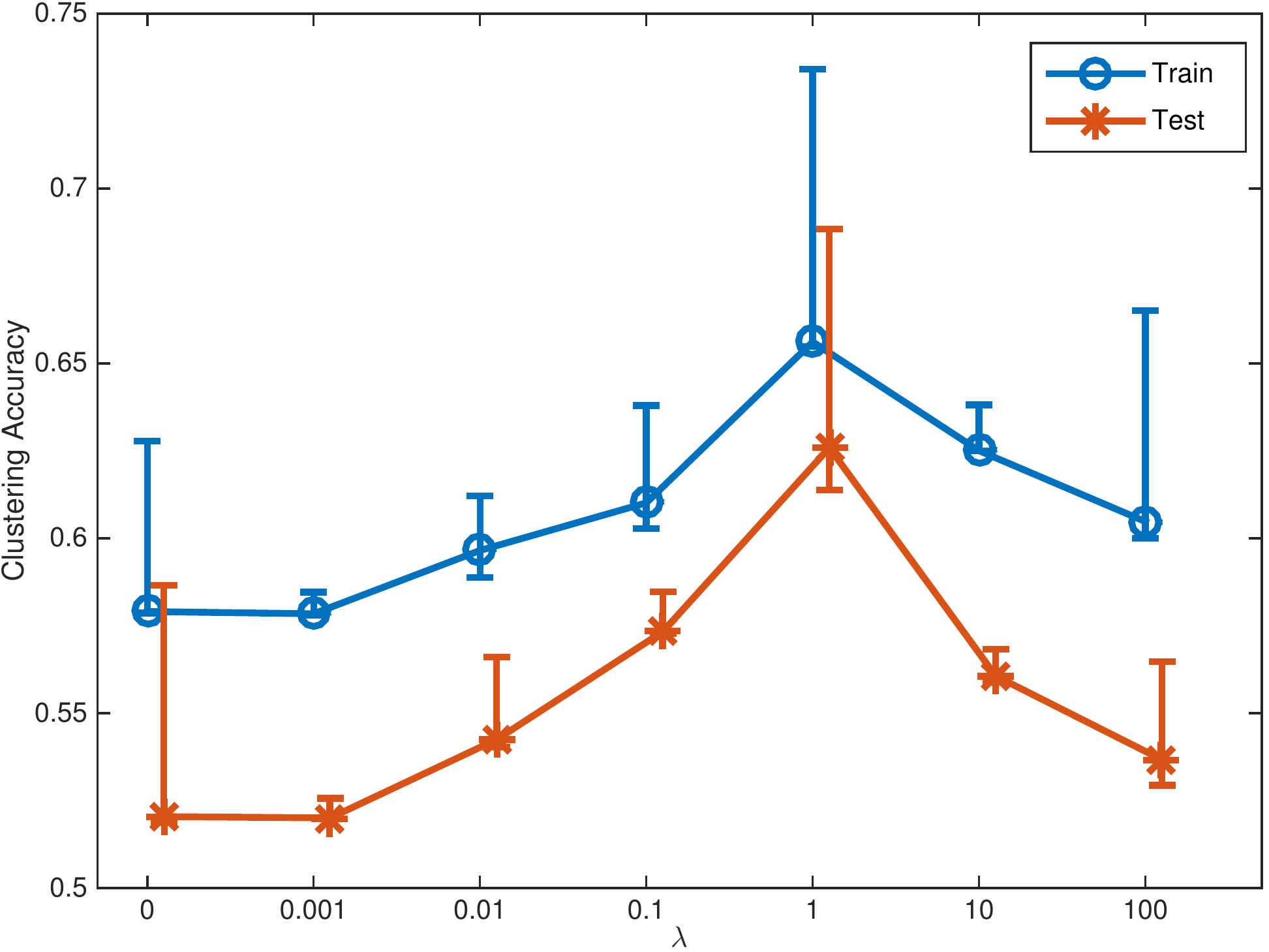}
\caption{Cora}
\label{fig:lambdacora}
\end{subfigure}
\caption{Effectiveness of diversity-encouraging prior for real-world datasets. $\lambda=0$ is the baseline of standard SePCA-MM.}
\label{fig:lambdaRealworld}
\end{figure*}

\begin{figure*}
\centering
\begin{subfigure}[b]{0.48\textwidth}
\includegraphics[width=\textwidth]{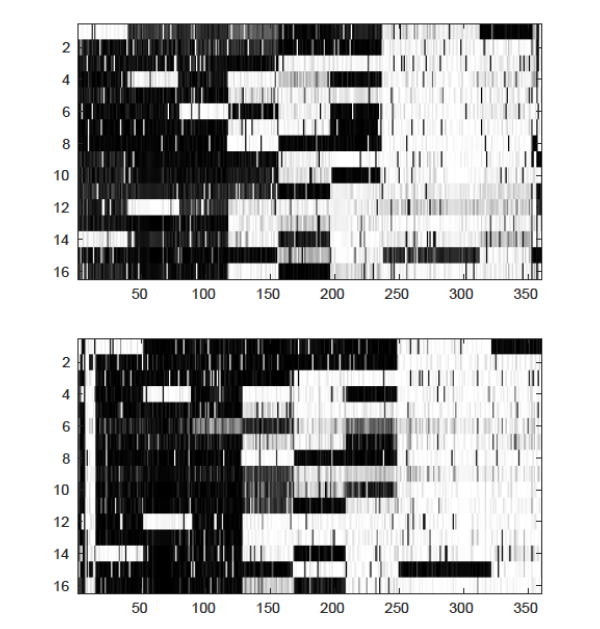}
\caption{Reconstructed mean parameters}
\end{subfigure}
\begin{subfigure}[b]{0.48\textwidth}
\includegraphics[width=\textwidth]{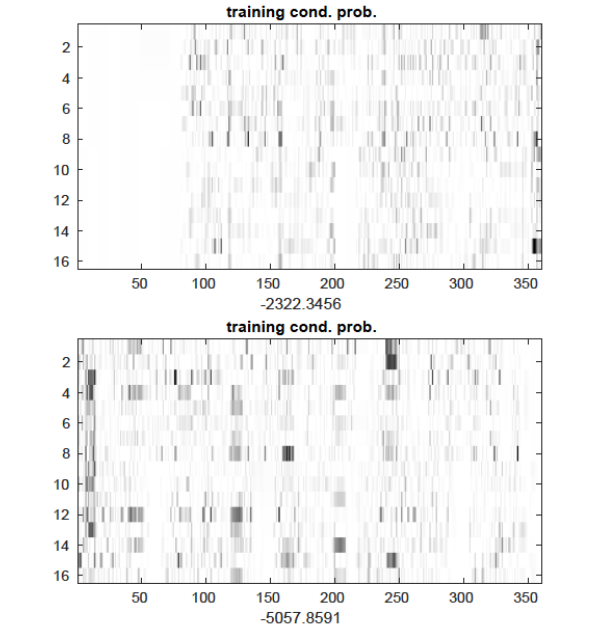}
\caption{Log-likelihood of one training dataset}
\end{subfigure}
\caption{The first row shows the results from traditional SePCA-MM, while the second row shows the results from the proposed DEPCAM with $d=6,K=3$. {(a) reconstructed mean parameters, i.e., $\alpha$. (b) log-likelihood of one training dataset given the reconstructed mean parameters (white indicates $0$ while darker color indicates larger negative value).}}
\label{fig:synPhi}
\end{figure*}

\subsection{Bernoulli Distributions}

One simple example from exponential family is the Bernoulli distribution, and is designated to handle binary-typed data sets.
We apply this kind of distribution throughout our experimental section. To make our paper self-contained, we briefly instantiate the general terms in our model, e.g., $g(WY)$ and $g'(WY)$, with the Bernoulli distribution.

The Bernoulli distribution with parameter $\alpha$ is formulated as $p(x;\alpha) =  \alpha^x (1-\alpha)^{1-x}$ and its corresponding natural parameters are  $\theta = \log \frac{\alpha}{1-\alpha}, g(\theta)=-\log (1+e^\theta)$, $h(x)=0$, and $\alpha = \frac{e^{\theta}}{e^{\theta}+1}$. Therefore,
\begin{align}
g(W^{z_n} y_n) &= \sum_{i=1}^D g([W^{z_n} y_n]_i), \\
g'(W^{z_n} y_n) &=
\begin{bmatrix}
\frac{1}{1+e^{[W^{z_n}y_n]_1}} -1 \\
\vdots \\
\frac{1}{1+e^{[W^{z_n}y_n]_D}} -1
\end{bmatrix}.
\label{eq:gderivative}
\end{align}

\section{Experimental Results}
\label{sec:exp}

\subsection{Synthetic Datasets}

The synthetic datasets are generated in a similar way with the one in \cite{mohamed2009bayesian}.
The reason we did not directly use their datasets is that they are used to verify a single PCA component, rather than a mixture models of PCAs. Therefore, we developed our own complex dataset in three steps.
First, three Bernoulli distributions with mean parameters $0.9,~0.5$, and $0.1$ were used to generate three categories of binary data, each of which contains three $16$-D binary prototype vectors with each bit independently drawn from $\{0,1\}$ with the same Bernoulli distribution.
Then, fifty duplicate copies of each prototype were made, composing a dataset with $450$ samples in total.
One data sample generated from these steps is illustrated in {Figure \ref{fig:trueD}}.
Finally, each bit was flipped with probability $0.1$ to generate a noisy sample as training dataset. One such sample is shown in Figure \ref{fig:noisyD}.

{The setting $\xi=0.1$, and $\varrho=0.1$ achieves best average predicting accuracy amongst $\{0.01, 0.1, 1, 10\}$ by performing 5-fold cross validation.}
All averaged experimental results are obtained via 5-fold split of the dataset.

\textbf{Diversity Effectiveness and model generalization ability:}
The effectiveness of the diversity-encouraging prior is explored by varying its weight $\lambda$ while keeping other parameters fixed, i.e., $K=3, d=4, \xi=0.1, \text{and } \varrho=0.1$.
The results of clustering accuracies are shown in Figure \ref{fig:lamError}. It is easily seen that with small to medium weight of diversity, the proposed method achieves significant improvement over the standard SePCA-MM in terms of clustering accuracies on both training and testing datasets.
Specially, the improvement on testing datasets demonstrates better generalization ability of the proposed model.
However, when the diversity is overemphasized in comparison to likelihood in mixture models, its performance dramatically deteriorates, as shown when $\lambda>10$.

\begin{figure*}
\centering
\begin{subfigure}[b]{0.45\textwidth}
\includegraphics[height=0.45\textwidth]{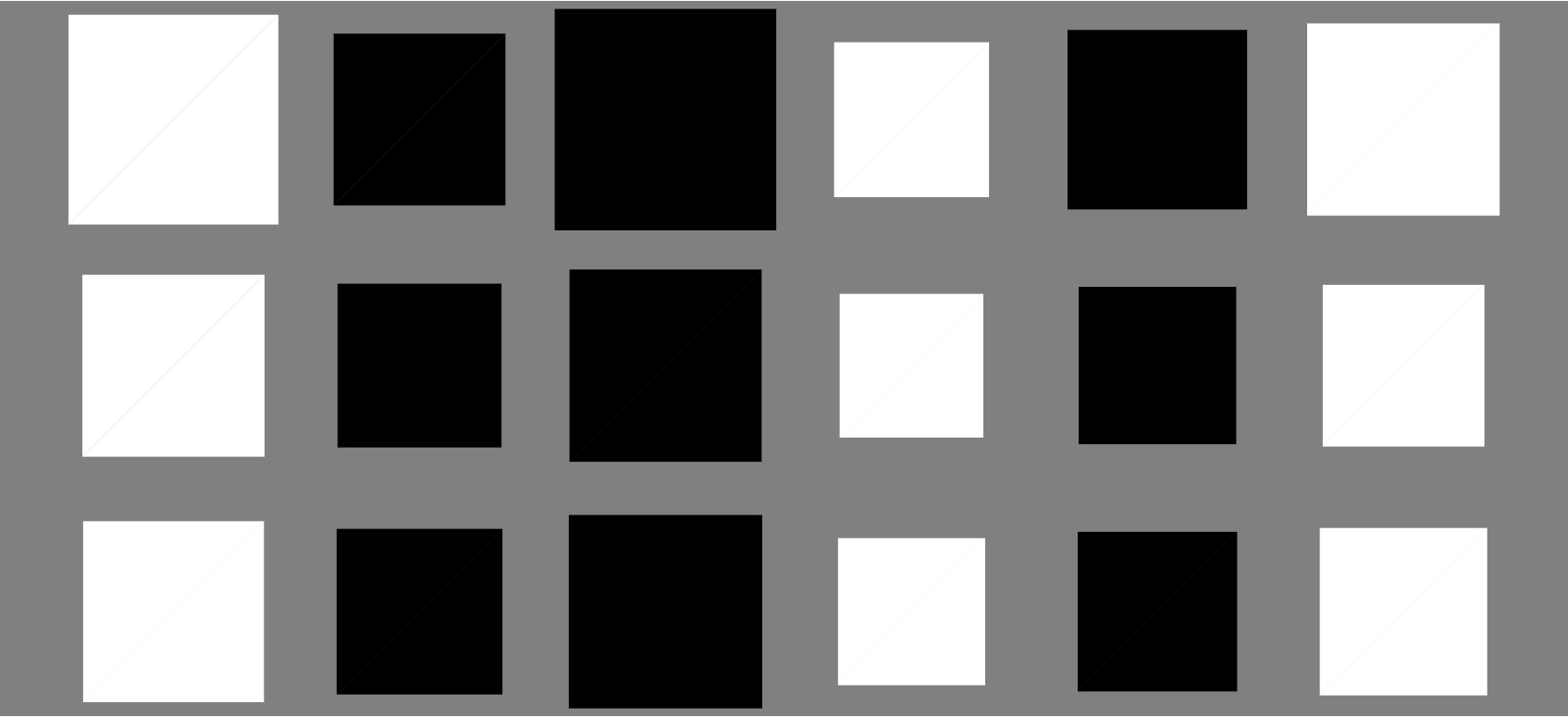}
\caption{Traditional SePCA-MM}
\label{fig:hinton0}
\end{subfigure}
\hspace{2em}
\begin{subfigure}[b]{0.45\textwidth}
\includegraphics[height=0.45\textwidth]{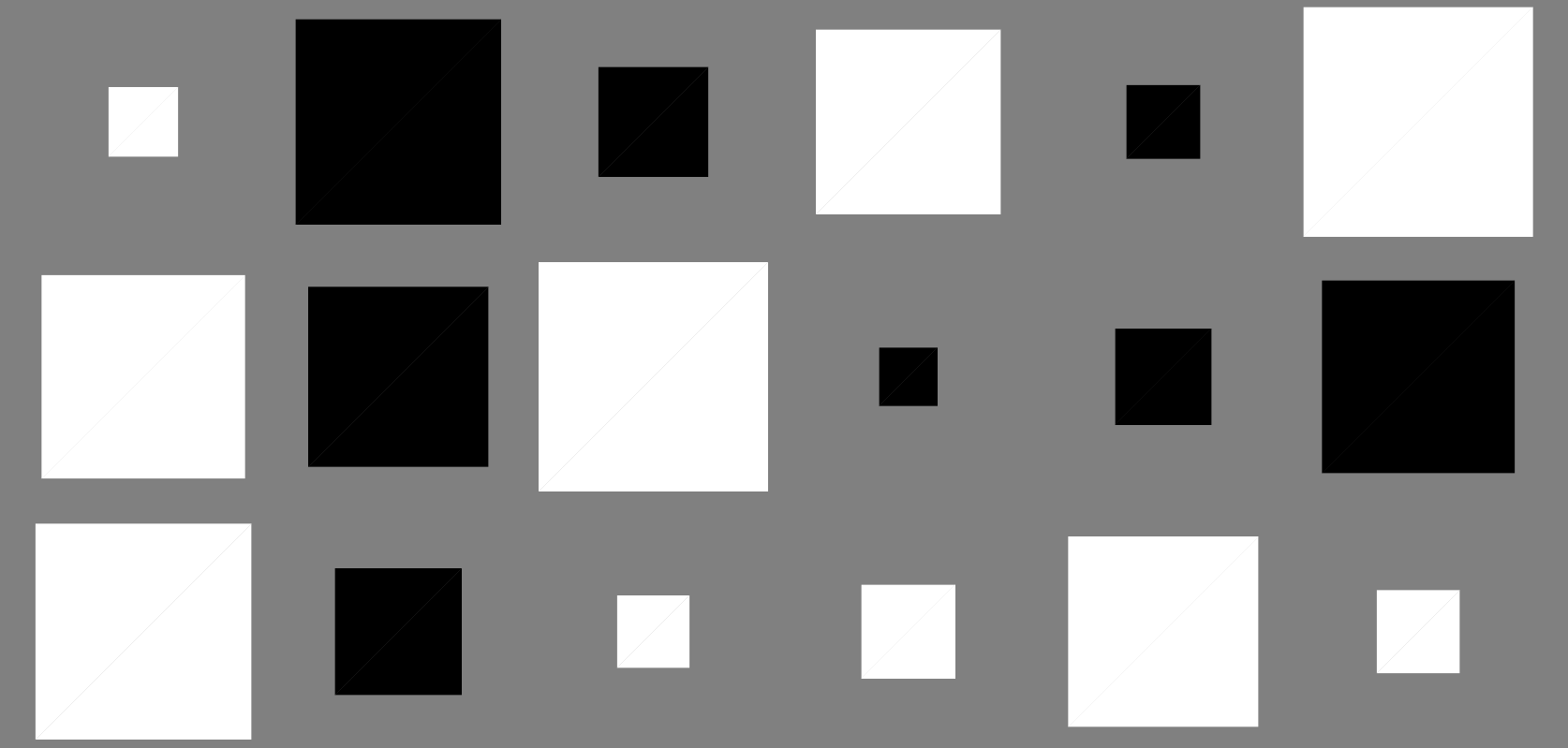}
\caption{DEPCAM}
\label{fig:hinton1}
\end{subfigure}
\caption{Hinton diagram of diagonal of $\{\Phi^k\}_{k=1}^K$, where white boxes indicating their positive elements, black ones indicating their negative ones, and the size of boxes symbolizing the elements' magnitude. }
\label{fig:synPhiHinton}
\end{figure*}

\begin{figure*}[!t]
\centering
\begin{subfigure}[b]{0.32\textwidth}
\includegraphics[width=\textwidth]{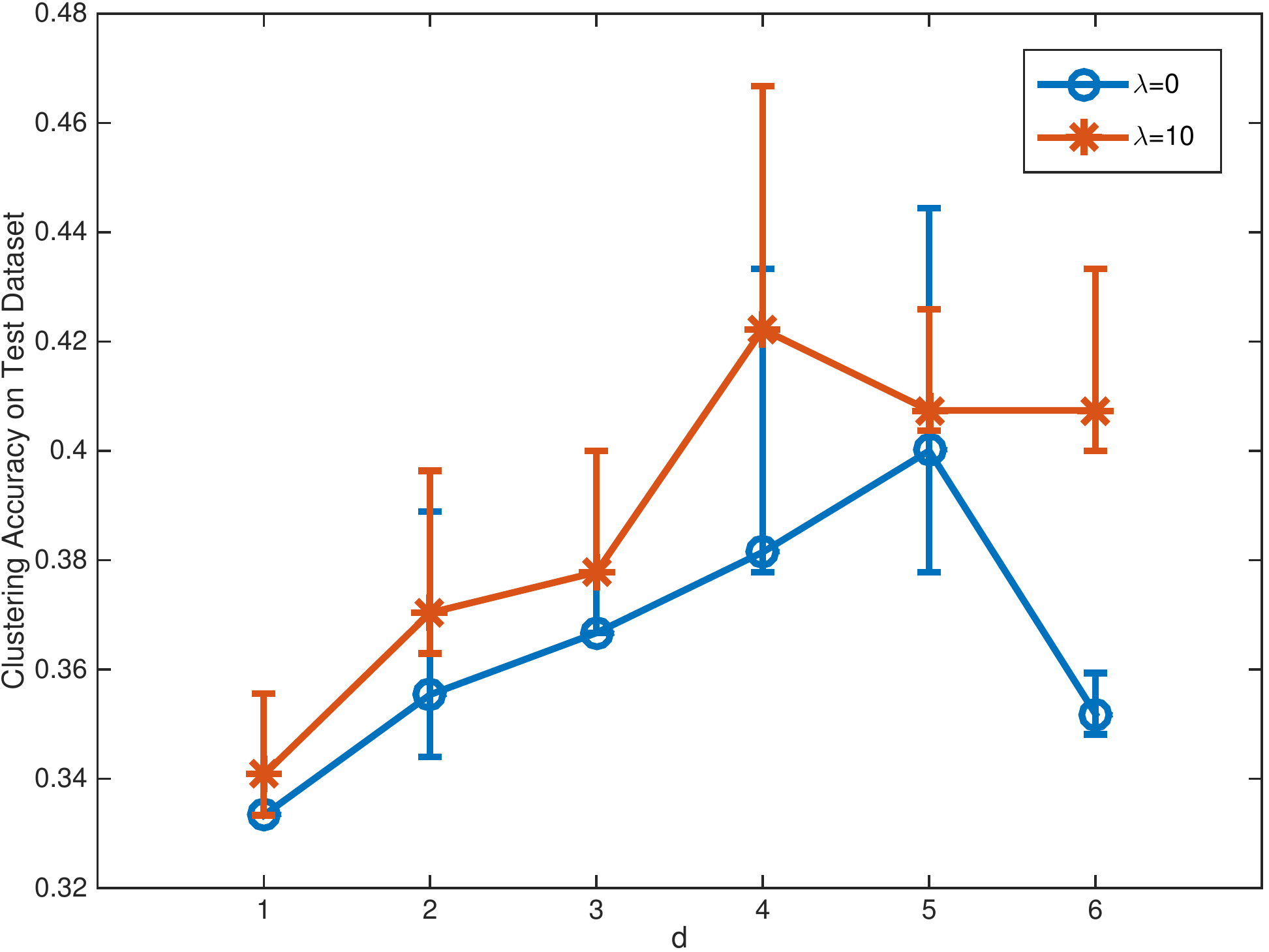}
\caption{Synthetic}
\label{fig:dSyn}
\end{subfigure}
\begin{subfigure}[b]{0.32\textwidth}
\includegraphics[width=\textwidth]{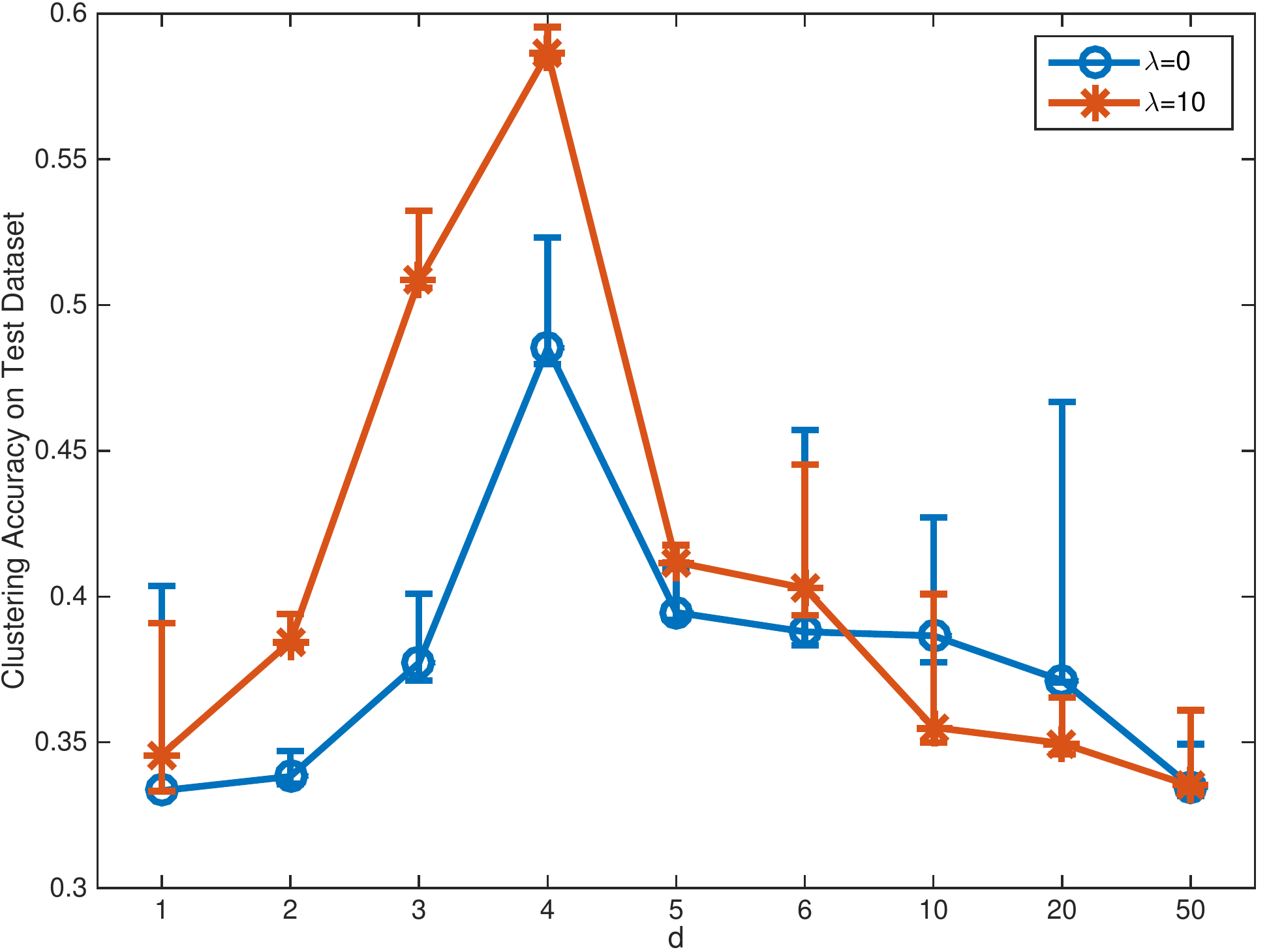}
\caption{USPS}
\label{fig:dUSPS}
\end{subfigure}
\begin{subfigure}[b]{0.32\textwidth}
\includegraphics[width=\textwidth]{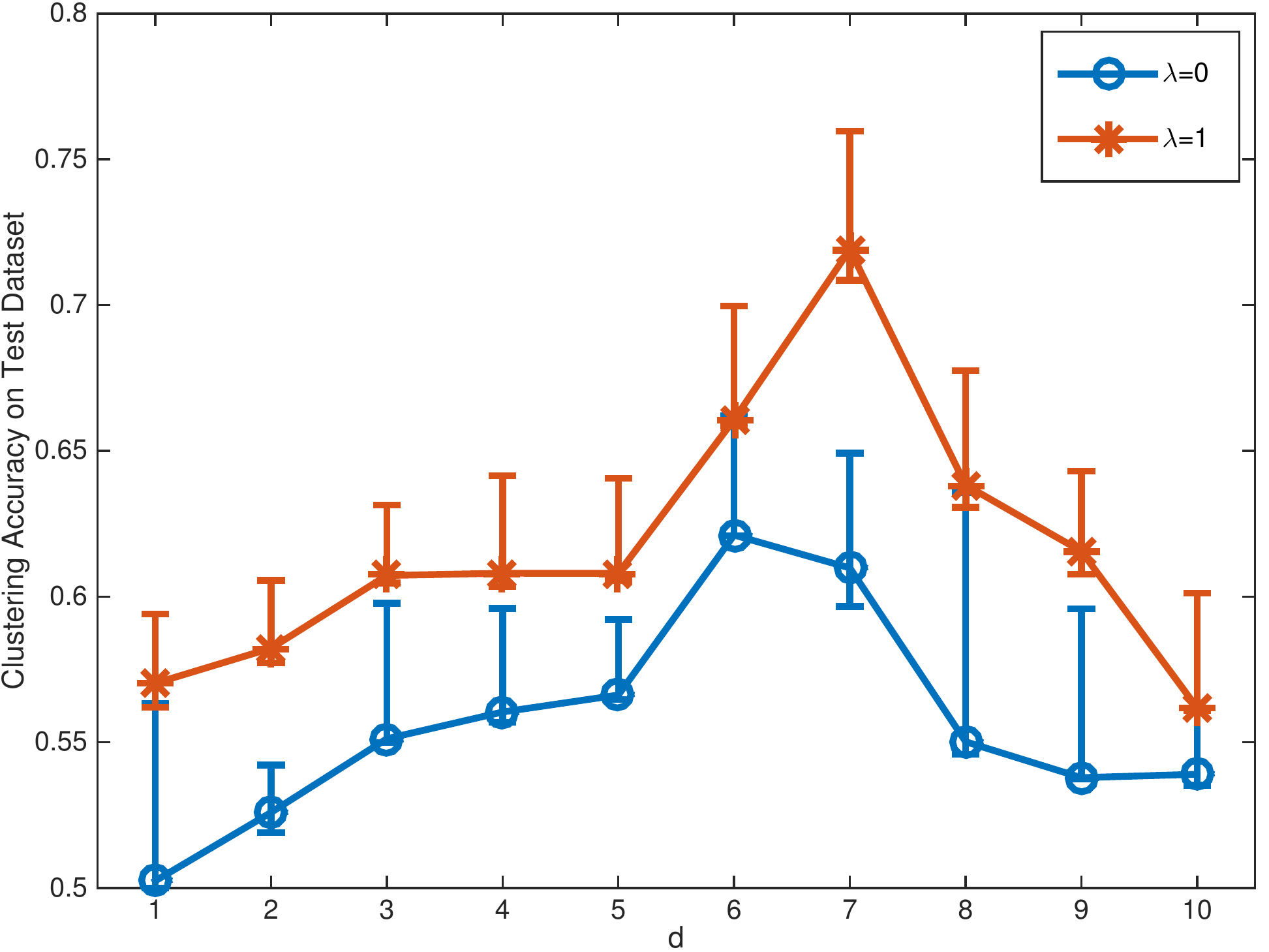}
\caption{Cora}
\label{fig:dcora}
\end{subfigure}
\caption{Model performance with varying $d$s, with one ($\lambda=0$) for SePCA-MM and the other for DEPCAM.}
\label{fig:lambdaRealworld}
\end{figure*}

\vspace{0.2em} \noindent
\textbf{Sparsity from $\ell_1$:}
The sparseness of $\{\Phi^k\}s$ with $d=6$ is examined by
comparing one diversified version ($\lambda=10$) to standard SePCA-MM ($\lambda=0$). The other parameters are the same as above.
The reconstructed mean parameters and likelihoods of one training dataset ({$360$ out of $450$ samples}) are shown in Figure \ref{fig:synPhi}.
On one hand, both of them achieve adequate results as shown in the left column. On the other hand, the baseline model SePCA-MM accomplishes slightly better results than the proposed diversified version in terms of likelihood,
which might due to its full employment of low-dimensional hidden spaces of each mixing component. The relevant $\{\Phi^k\}s$ are visualized with Hinton diagram and shown in Figure \ref{fig:synPhiHinton}.
{
In Figure \ref{fig:hinton0}, the value distribution of $\{\Phi^k\}$ for the three different components are quite similar. In addition, they choose all dimensions in the hidden space as their effective dimensions, as indicated by the similar but large sizes of all the boxes. By contrast, the ones learned by the proposed model exhibit unique properties as shown in Figure \ref{fig:hinton1}. First, the value distributions for different components vary. Second, the effective dimensions are less than the ones needed by traditional model, as indicated by several quite small boxes which could be ignored for reconstruction.
}
We amount all these properties to the $\ell_1$ prior, which forces $\{\Phi_k\}s$ to be sparse.
Such sparseness provides the mixture model with the power of automatically determining the dominant PCs of each mixing component. In addition, it reduces model complexity in a certain degree.

 \vspace{0.2em}\noindent
\textbf{Induced model parsimony:}
The model parsimony induced by the proposed diversified SePCA-MM is further verified by varying dimensions of hidden spaces, namely, parameter $d$. The other parameters are kept the same as above and $\lambda$ is varied from $\{0,10\}$ for comparison.
The performance 
in terms of averaged clustering accuracy on hold-out testing datasets
are shown in Figure \ref{fig:dSyn}. Accordingly, DEPCAM can achieve better performance with similar or even smaller $d$.
For instance, the proposed model  with $d=3$ achieves as high accuracy as the baseline with higher hidden space dimension of $d=4$. Another example is that with $d=4$, DEPCAM achieves the best clustering accuracy which is significantly higher than any achieved by baseline with different $d$s.
The decline along with increment of $d$ for both methods might be caused by encoding more noise.

\subsection{Real-world Datasets}
We tested our model by conducting experiments on two real-world datasets, i.e., the USPS hand-written digits and the Cora dataset\footnote{http://linqs.umiacs.umd.edu/projects/projects/lbc/}.
Part of these datasets are used to train the diversified mixture models.
For USPS, as the same with \cite{li2013simple}, images from three digit categories of $2$, $3$, and $4$ compose our dataset. Two hundred images of each category are randomly selected.
%
For Cora, four hundred papers were randomly selected from two categories: Case-Based and Genetic-Algorithms.
Five-fold  cross validation is performed to estimate clustering accuracies for both datasets.

\vspace{0.2em}
\noindent \textbf{Diversity Effectiveness:}
{The diversity effectiveness of the proposed DEPCAM is explored by varying the diversity-encouraging weight $\lambda$ while fixing other parameters, which are obtained by grid searching:
For USPS, $d=4,~K=3,~\xi=0.005,~\varrho=0.1$; For Cora, $d=7,~K=2,~\xi=0.001,~\varrho=0.1$.}
The results are shown in Figure \ref{fig:lambdaUSPS} and Figure \ref{fig:lambdacora}.
For both datasets, DEPCAM with a medium sized $\lambda$ achieves the best clustering accuracies on both training and testing datasets. It evidences the effectiveness of diversity prior.

\noindent \textbf{Induced model parsimony:}
The held-out clustering accuracies with varying $d$ are shown
in Figure \ref{fig:dUSPS} and Figure \ref{fig:dcora}.
It shows that to achieve equal clustering accuracies, DEPCAM consistently
needs smaller $d$s. To some extend, it proves DEPCAM induces more parsimonious models than
the standard SePCA-MM.

\begin{figure}[!t]
\centering
\includegraphics[width=0.5\textwidth]{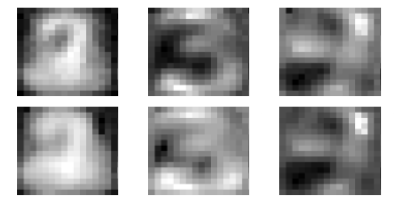}
\caption{The fitted three local PCA transformation matrices from the proposed DEPCAM with $K=3, d=2$ for USPS digits dataset. Each row represents one PC. }
\label{fig:transMatrixUSPS}
\end{figure}

\vspace{0.2em} \noindent \textbf{Qualitative results:}
Two dimensions with larger qualities of $\Phi$ among the four, of one learned three PCA mixing components, are shown in Figure \ref{fig:transMatrixUSPS}. It is trained with parameters $K=3, d=4, \lambda=10, \xi=0.005, \rho=10$.
 It can be easily seen that the learned dominant PCs of the proposed diversified model are able to capture the main shape for each digit.

\begin{figure}[!t]
\centering
\begin{subfigure}[b]{0.5\textwidth}
\includegraphics[width=\textwidth]{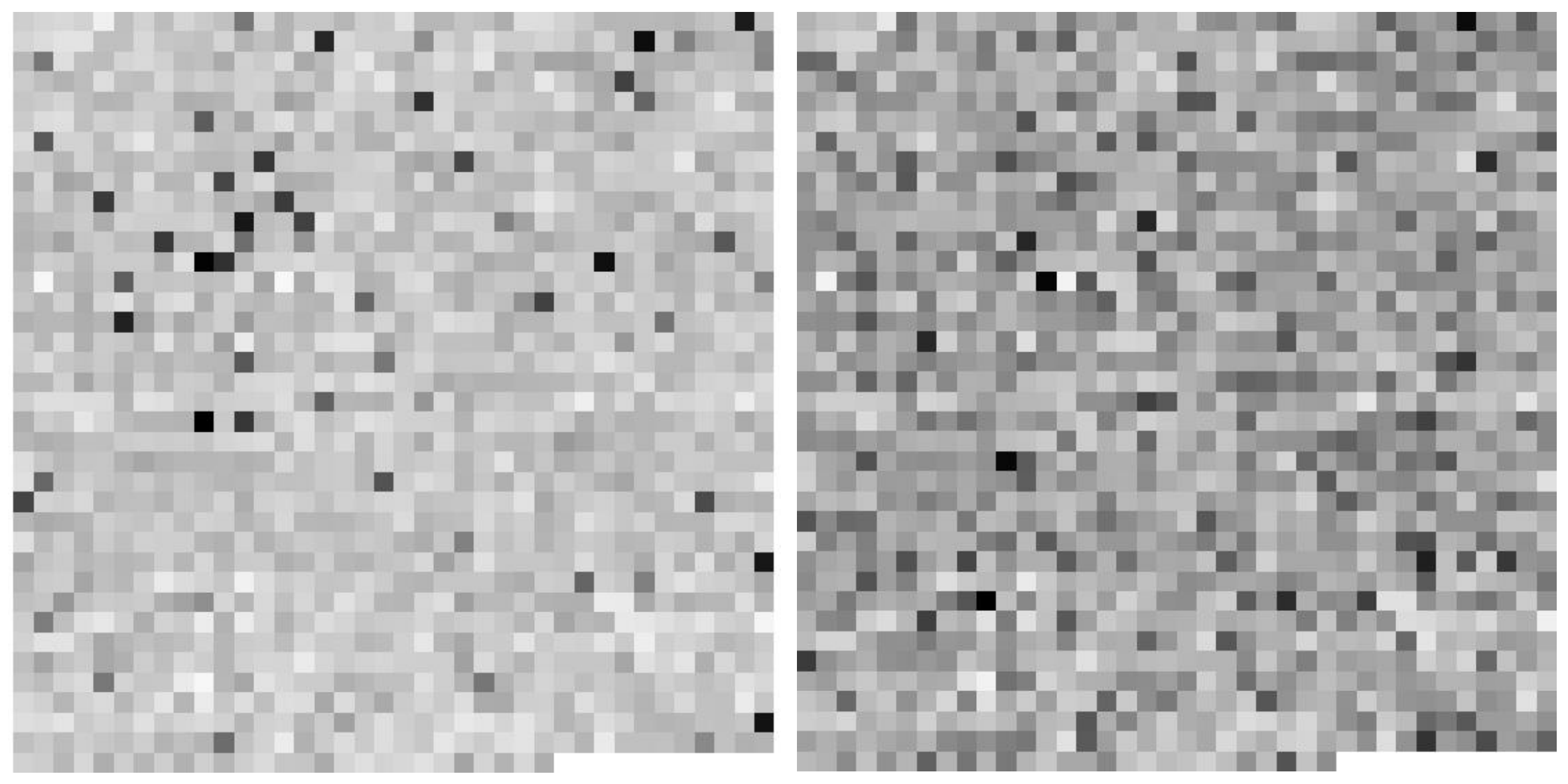}
\caption{SePCA-MM}
\end{subfigure}
\begin{subfigure}[b]{0.5\textwidth}
\includegraphics[width=\textwidth]{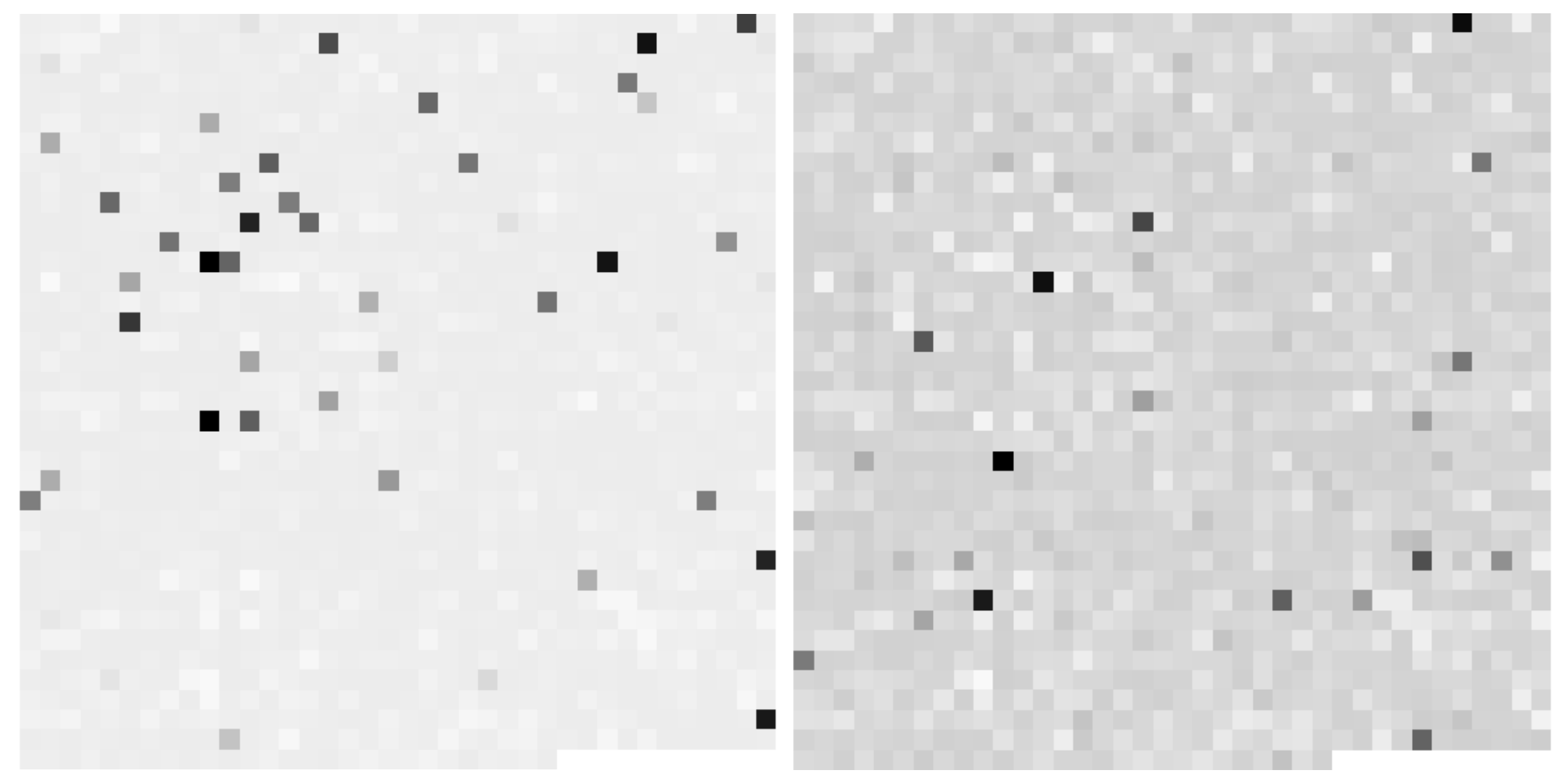}
\caption{DEPCAM}
\end{subfigure}
\caption{Dominant PCs for Cora datasets (Dark black pixels indicating larger absolute values than light grey pixels).}
\label{fig:transMatrixCora}
\end{figure}

To demonstrate the sparsity, dominant PCs of two transformation matrices learned for Cora dataset from both standard SePCA-MM and the diversified setting are shown in Figure \ref{fig:transMatrixCora}.
It is obvious that most of values obtained by the proposed DEPCAM are zero approaching. In other words, the PCs are much sparser than the one obtained by the standard mixture model.

\section{Conclusion}
\label{sec:conclusion}

In this paper we propose a diversified exponential family PCA mixture (DEPCAM) model.
It effectively introduces a diversity-encouraging prior to mixing components of standard SePCA-MM.
The proposed model is considered to be effective due to three main advantages.
First, the explicit repulsiveness among mixing components reduces model redundancy and the probability of overlapping. Second, the delicately designed diversity-encouraging prior, specifically the decomposition of transformation matrices into quality features and diversity features, provides a straightforward way to determine the dominating PCs of each mixing component. In addition, sparseness of quality terms is introduced with $\ell_1$ constraints.
Empirical results verify the effectiveness of the proposed model. Our future work will focus on more accurate approximations, such as Monte Carlo methods and variational inference methods.


\ifCLASSOPTIONcaptionsoff
  \newpage
\fi

\bibliographystyle{IEEEtran}

\ifCLASSOPTIONcaptionsoff
  \newpage
\fi
\bibliography{bibtex}

\begin{IEEEbiography}[{\includegraphics[width=1in,height=1.25in,clip,keepaspectratio]{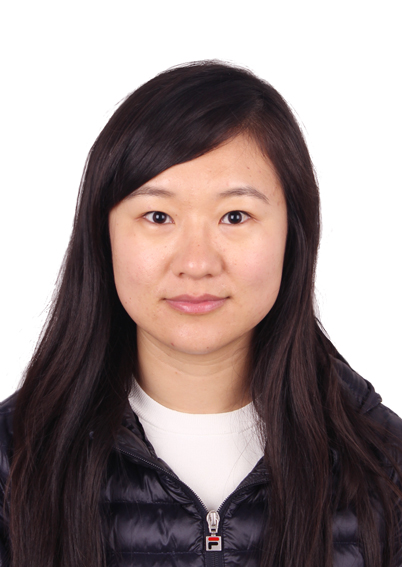}}]{Maoying Qiao}
received the B.Eng. degree in Information Science and Engineering from Central South University, Changsha, China, in 2009, and the M.Eng. degree in Computer Science from Shenzhen Institutes of Advanced Technology, Chinese Academy of Sciences, Shenzhen, China, in 2012. She is currently working towards the PhD degree in the University of Technology Sydney.
Her current research topics are pattern recognition and probabilistic graphical modeling.
\end{IEEEbiography}
\vspace{-3em}

\begin{IEEEbiography}[{\includegraphics[width=1in,height=1.25in,clip,keepaspectratio]{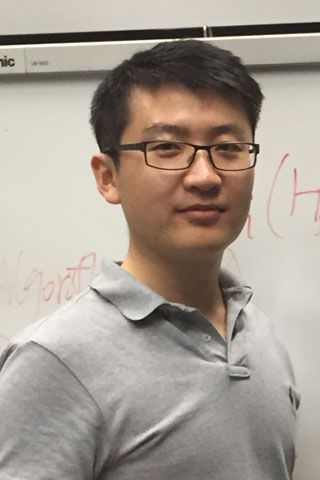}}]{Tongliang Liu}
received the BE degree in Electronic Engineering and Information Science from the University of Science and Technology of China, Hefei, China, in 2012, and the PhD degree from the University of Technology Sydney, Sydney, Australia, in 2016. He is currently a Lecturer with the Faculty of Engineering and Information Technology in the University of Technology Sydney. His research interests include statistical learning theory, computer vision, and optimization. He has authored and co-authored 10+ research papers including IEEE T-PAMI, T-NNLS, T-IP, NECO, ICML, KDD, IJCAI, and AAAI. He won the Best Paper Award in IEEE International Conference on Information Science and Technology 2014.
\end{IEEEbiography}
\vspace{-3em}

\begin{IEEEbiography}[{\includegraphics[width=1in,height=1.25in,clip,keepaspectratio]{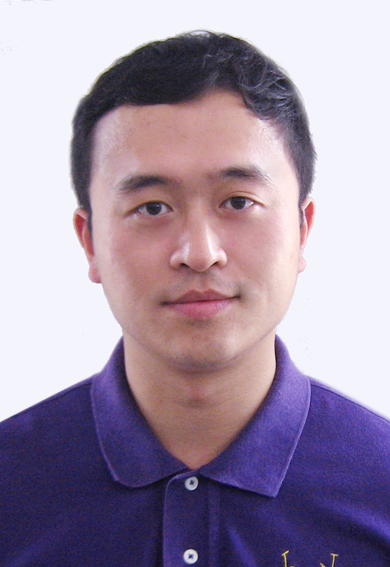}}]{Jun Yu}
(M'13) received the B.Eng. and PhD degrees from Zhejiang University, Zhejiang, China.
He is currently Professor with the School of Computer Science and Technology, Hangzhou
Dianzi University, Hangzhou, China. He was Associate Professor with the School of
Information Science and Technology, Xiamen University, Xiamen, China. From 2009 to 2011,
he was with Singapore Nanyang Technological University, Singapore. From 2012 to 2013, he was a
Visiting Researcher with Microsoft Research Asia, Beijing, China. He has authored and co-authored over 80 scientific articles.
His current research interests include multimedia analysis, machine learning,
and image processing. Prof. Yu was the Co-Chair for several special sessions, invited sessions, and workshops. He served as a Program Committee Member or a Reviewer for top conferences and prestigious journals. He is a Professional Member of ACM and China Computer Federation.
\end{IEEEbiography}

\begin{IEEEbiography}[{\includegraphics[width=1in,height=1.25in,clip,keepaspectratio]{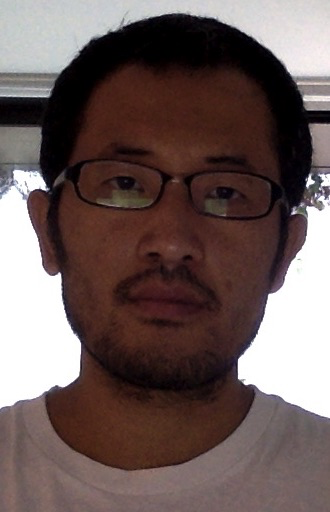}}]{Wei Bian}
(M'14) received the B.Eng. degree in electronic engineering and the B.Sc. degree in applied mathematics in 2005, the M.Eng. degree in electronic engineering in 2007, all from the Harbin institute of Technology, harbin, China, and the PhD degree in computer science in 2012 from the University of Technology, Sydney.
His research interests are pattern recognition and machine learning.
\end{IEEEbiography}
\begin{IEEEbiography}[{\includegraphics[width=1in,height=1.25in,clip,keepaspectratio]{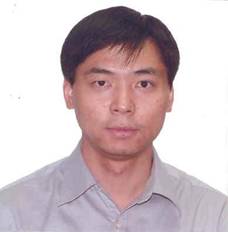}}]{Dacheng Tao}
(F'15) is Professor of Computer Science and Director of the Centre for Artificial Intelligence, and the Faculty of Engineering and Information Technology in the University of Technology Sydney. He mainly applies statistics and mathematics to Artificial Intelligence and Data Science. His research interests spread across computer vision, data science, image processing, machine learning, and video surveillance. His research results have expounded in one monograph and 200+ publications at prestigious journals and prominent conferences, such as IEEE T-PAMI, T-NNLS, T-IP, JMLR, IJCV, NIPS, ICML, CVPR, ICCV, ECCV, AISTATS, ICDM; and ACM SIGKDD, with several best paper awards, such as the best theory/algorithm paper runner up award in IEEE ICDM'07, the best student paper award in IEEE ICDM'13, and the 2014 ICDM 10-year highest-impact paper award. He received the 2015 Australian Scopus-Eureka Prize, the 2015 ACS Gold Disruptor Award and the 2015 UTS Vice-Chancellor's Medal for Exceptional Research. He is a Fellow of the IEEE, OSA, IAPR and SPIE.
\end{IEEEbiography}

\end{document}